\documentclass[authoryear, 12pt]{elsarticle}

\usepackage{amssymb}

\usepackage{hyperref}
\usepackage{booktabs}
\usepackage{multirow}
\usepackage{caption}
\usepackage{subcaption}
\usepackage{algorithm}
\usepackage{algpseudocode}

\makeatletter
\providecommand{\doi}[1]{%
  \begingroup
    \let\bibinfo\@secondoftwo
    \urlstyle{rm}%
    \href{http://dx.doi.org/#1}{%
      doi:\discretionary{}{}{}%
      \nolinkurl{#1}%
    }%
  \endgroup
}
\makeatother

\usepackage{amsmath}
\newcommand{\vect}[1]{\boldsymbol{\mathbf{#1}}}

\DeclareMathOperator*{\argmax}{arg\,max}

\newcommand{\nOSmeasurements}{$180$~}

\journal{Sustainable Energy, Grids and Networks}

\begin{document}

\begin{frontmatter}



\title{Acquiring Better Load Estimates by Combining Anomaly and Change Point Detection in Power Grid Time Series Measurements}


\author[inst1]{Roel Bouman}
\author[inst2]{Linda Schmeitz}
\author[inst1]{Luco Buise}

\author[inst2]{Jacco Heres}
\author[inst1]{Yuliya Shapovalova}
\author[inst1]{Tom Heskes}

\affiliation[inst1]{organization={Institute for Computing and Information Sciences, department of Data Science, Radboud University},
            addressline={Toernooiveld 212}, 
            city={Nijmegen},
            postcode={6525EC}, 
            country={The Netherlands}}

\affiliation[inst2]{organization={Advanced Analytics, Alliander N.V.},
            addressline={Utrechtseweg 68}, 
            city={Arnhem},
            postcode={6812AH}, 
            country={The Netherlands}}

\begin{abstract}
In this paper we present novel methodology for automatic anomaly and switch event filtering to improve load estimation in power grid systems. By leveraging unsupervised methods with supervised optimization, our approach prioritizes interpretability while ensuring robust and generalizable performance on unseen data. Through experimentation, a combination of binary segmentation for change point detection and statistical process control for anomaly detection emerges as the most effective strategy, specifically when ensembled in a novel sequential manner. Results indicate the clear wasted potential when filtering is not applied. The automatic load estimation is also fairly accurate, with approximately 90\% of estimates falling within a 10\% error margin, with only a single significant failure in both the minimum and maximum load estimates across 60 measurements in the test set. Our methodology's interpretability makes it particularly suitable for critical infrastructure planning, thereby enhancing decision-making processes. 
\end{abstract}




\end{frontmatter}


\section{Introduction}
\label{introduction}
The global energy landscape is undergoing a transformative shift towards sustainability, driven by the urgent need to mitigate climate change and reduce reliance on fossil fuels~\citep{united2021theme}. This process, more commonly known as the energy transition, presents a multitude of challenges that must be addressed to achieve a successful energy transition. These challenges encompass technical, economic, social, and political aspects, demanding innovative solutions and collaborative efforts on a global scale~\citep{laes2014comparison}. One of the key bottlenecks in implementing the energy transition in the Netherlands is the growth of electrical infrastructure~\citep{zuijderduin2012integration}. In order to replace fossil fuel energy sources, the capacity of the power grid needs to increase significantly. However, increasing the capacity of the power grid involves several challenges. Due to a scarcity of resources~\citep{moss2011critical} identification of key areas where additional capacity is most needed is imperative. 

In addition to the need for additional capacity, the way the grid is being used is also changing~\citep{brouwer2013fulfilling}. Due to the increasing reliance on solar and wind energy, the centralized production of energy is a fading paradigm. Decentralized production in the form of a multitude of wind and solar parks, as well as solar panels covering a large percentage of urban housing, are changing the ways in which electricity is distributed~\citep{van2006optimal, pagani2011towards}. Where previously gas was the primary source of heating in houses, heat pumps are increasingly being used in households~\citep{kieft2021heat}. Electrical cooling of houses has seen a substantial increase in recent years~\citep{hekkenberg2009indications}. Electric motors are quickly replacing fossil fuel combustion engines in personal vehicles~\citep{wang2021grid}. These vehicles are now often charged at home, at the workplace, or near other hubs~\citep{funke2019much}. These changing requirements drive a need to expand the power grid in a smart manner, where expansion is done there where it is most needed in the near future.

Next to expansion, smarter use of the existing grid is essential. This can be done through for example flexible energy contracts, the use of redundant grid capacity for power generation, and the use of batteries~\citep{sufyan2019sizing}. Better insight into grid usage over time is needed to facilitate these changes. With more measurements this can be achieved, but this means also more data that needs to be cleaned before it can be used for analysis.

\begin{figure}
\includegraphics[width=\textwidth]{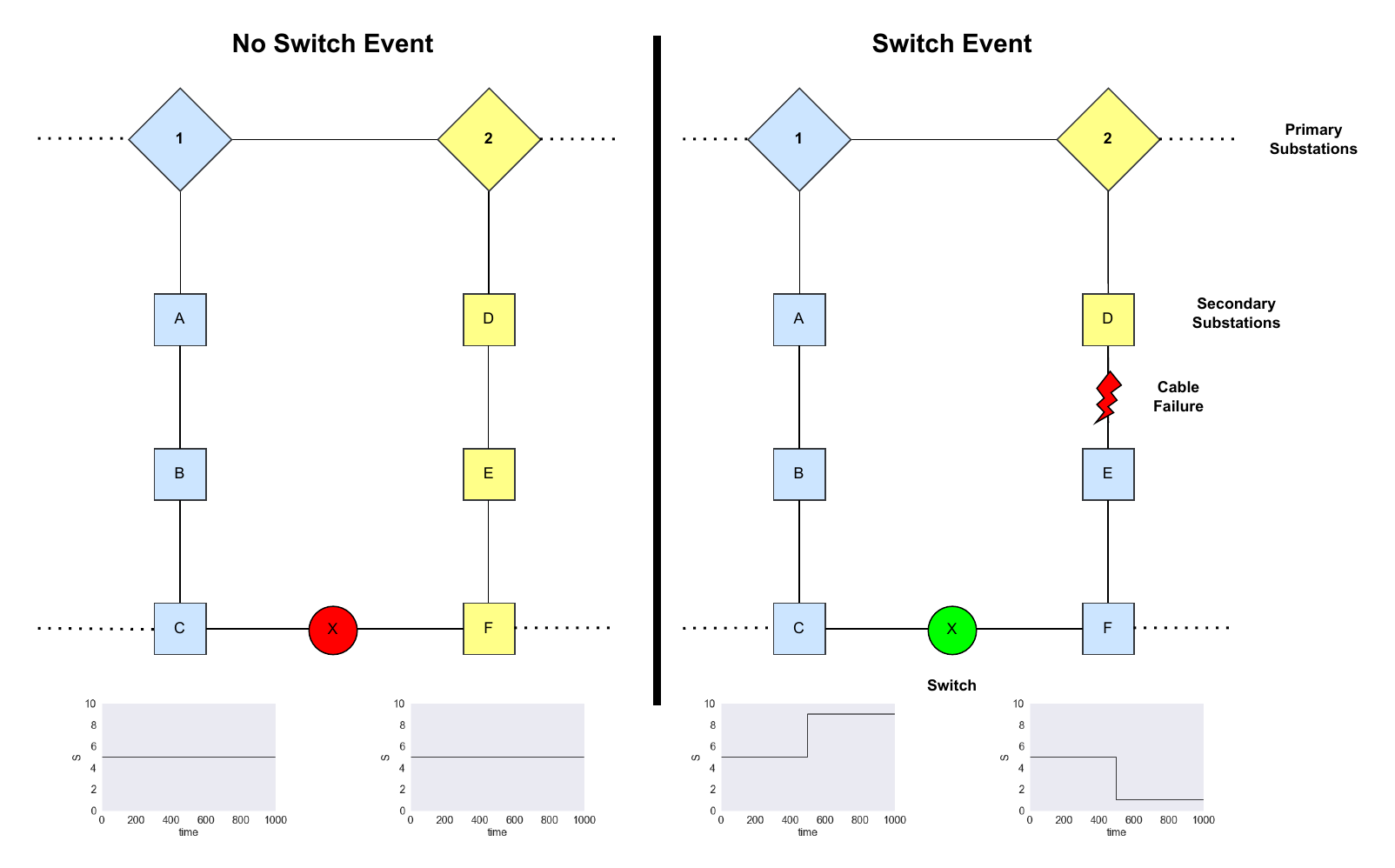}
\caption{An illustration of a switch event. The diamonds numbered 1 and 2 indicate two primary substations. The squares named A-F indicate secondary substations. The circle named X indicates a switch. Solid lines indicates connections between substations, and dashed lines indicate connections to other substations outside of the figure. The primary substation and the secondary substations powered by it are colored light blue or light yellow for primary substations 1 and 2 respectively. When a switch event occurs, for example due to cable failure between secondary substations D and E, power will be supplied from primary substation 1 rather than 2. This is indicated by the switch X color change from red to green. This leads to a temporary increase in apparent power measured at 1 and a decrease at 2, as illustrated with the apparent power measurements as a function of time at the bottom of the figure. } 
\label{fig:switch_event}
\end{figure}

In order to determine key points of expansion, an accurate overview of the current state of grid capacity needs to be made. In this study, we specifically study primary substation-level measurements on the Dutch power grid managed by Alliander. In order to know what percentage of a primary substation's capacity is being used, accurate estimates of the minimum and maximum load of the subgrid that substation supplies must be made. This process is more commonly known as demand modeling~\citep{betge2021efficient}. Primary substation-level time series load measurements, however, cannot be directly interpreted without giving a distorted view. These measurements are contaminated with anomalies, for example caused by measurement errors, and with switch events. In a switch event power from different primary substations is rerouted to or from part of the measured primary substation's route. An example of this can be found in Figure~\ref{fig:switch_event}, where due to a cable failure power is rerouted from primary substation 1 to supply secondary substations E and F. 
Rerouting power reduces the blackout time in case of damage to a cable in the power grid, and is possible when the grid is sufficiently redundant to circumvent the broken cable. Switch events can also happen when long-term maintenance is performed on a primary substation or any its cables or secondary substations. Switch events can have a large range of possible lengths, from a few minutes, to multiple months, depending on how fast the underlying problem is resolved. Both of these, measurement errors, which we will call anomalies, and switch events, need to be filtered out to get an accurate estimate of the load profile. This more accurate load profile can be used directly for more optimized usage of the grid. It can then be used to find the true minimum and maximum load of the substation under normal operating conditions, so the redundancy capacity can be added separately. The task of detecting which parts of a load measurement represent the normal situation can be seen as a time series segmentation task. 

Traditionally, these load measurement are manually segmented and annotated by domain experts within the Alliander organization. This procedure is however extremely time consuming, and produces annotations with a high percentage of label noise. This label noise is of little concern in determining the minimum and maximum loads, as the annotation procedure is generally optimized in such a way that these loads are as accurate as possible. The resulting annotation can however not be used for other purposes such as year-round usage insights. An automated annotation procedure, capable of producing high-quality annotations, can therefore both free up valuable domain expert time, as well as allow for detailed studies on load measurement data previously unfeasible. Because economically and logistically vital decisions are made based on these load estimates, it is extremely important that any machine learning model automating this procedure is highly interpretable.

For smart grids, change point and anomaly detection algorithms have been broadly applied. 
\citet{zhang2021time} describe multiple time series anomaly detection algorithms used in various applications across smart grids. They differentiate between point, contextual, and pattern anomalies. 
\citet{thomas2020passive} describe a novel method tailored to detecting islanding. In addition to islanding, their method, which uses $k$-means clustering and empirical mode decomposition, is able to detect load switch events and other faults. 
\citet{neagoe2021change} used change point detection in order to detect multi-year patterns in hydropower generation in Romania. 
\citet{wang2020real} combine a regression-based anomaly detection model with SVM, $k$NN and a cross-entropy loss function to detect anomalies in a Tanzanian solar power plant.
\citet{wang2023real} present methodology for simultaneously performing load forecasting and semi-supervised anomaly detection and apply it to the UCI electric load dataset. They report higher specificity and sensitivity than neural network-based methods, but find similar F1 scores.
For shorter time series, \citet{rajabi2020comparative} have compared various clustering methods for load pattern segmentation. 

Load estimation has been performed across a variety of use cases within power grid operation and study. 
\citet{heslop2016maximum} have estimated maximum photovoltaic generation for residential low voltage feeders.
\citet{mendes2023signal} estimate the variability of the load, using graph signal processing, on data with a high level of distributed generation, similar to the situation in the Netherlands. 
\citet{langevin2023efficient} employ variational autoencoders for short-term forecasting of the load within households.
\citet{kara2018disaggregating} use multiple linear regression to disaggregate the solar generation in regular feeder measurements in order to get separate estimates for generation and production.
\citet{asefi2023anomaly} perform a combination of statistical modeling, anomaly detection and classification in order to estimates states and identify false data injection.

However, to the best of our knowledge, for time series, of a year or longer in length, where event lengths vary substantially, no studies have been conducted on load estimation through means of automated segmentation. 

\section{Materials and Methods}
\subsection{Data}
In this study we optimize and evaluate our algorithms for anomaly and switch event detection on a total of \nOSmeasurements primary substation load measurements. Most primary substation measurements span a full year in length, and provide measurements at regular 15-minute intervals of the apparent power $S$, which we will refer to as load. We calculate $S$ from the active power $P$ and the reactive power $Q$ and assign it the sign of $P$, $S = sign(P)*\sqrt{P^2+Q^2}$. In some cases, the measurement equipment of a primary substation does not allow for accurate measurements of both $P$ and $Q$. In these cases, $S$ is calculated from $S = \sqrt{3} * V * I$. The $\sqrt{3}$ term originates from the fact that in a 3-phase system the phase voltage rather than the line voltage is used.

At each 15-minute interval over which the load is measured, we also calculate the so-called bottom-up load $B$ throughout the subgrid supplied by that specific primary substation. The bottom-up load is an estimation of the load over a certain primary substation, but not a direct measurement, like $S$. Thus, the bottom-up load is directly related to the actual load measurement. The bottom-up is traditionally used by distribution system operators to get an estimate of load on the grid on places where no measurements are available. This bottom-up estimate tries to reconstruct the total load based on telemetry measurements from bulk consumers, from aggregated smaller scale measurements, and from average profiles based on smart meters at consumers' homes and some smaller bulk consumers. In the latter machine learning is used to estimate the load profiles of those consumers that do not have a smart meter, or have not consented to have their smart meter data read~\citep{heres2023creating}. In order to acquire the final bottom-up load, Alliander uses the SunDance algorithm~\citep{chen2017sundance} for disaggregation of net consumption and generation, $k$-means clustering~\citep{macqueen1967some, lloyd1982least} for generating load profiles, and XGBoost~\citep{chen2016xgboost} for determining which clustered load profiles should be used instead of missing smart meter measurements~\citep{heres2023creating}.  More details regarding the bottom-up generation methodology can be found in \citep{heres2023creating}. Most often, the bottom-up load is fairly accurate. Most failure cases are not caused by the algorithm, but rather by incorrect grid-topology data, causing consumers to be wrongfully included or excluded.

\begin{figure}
\includegraphics[width=\textwidth]{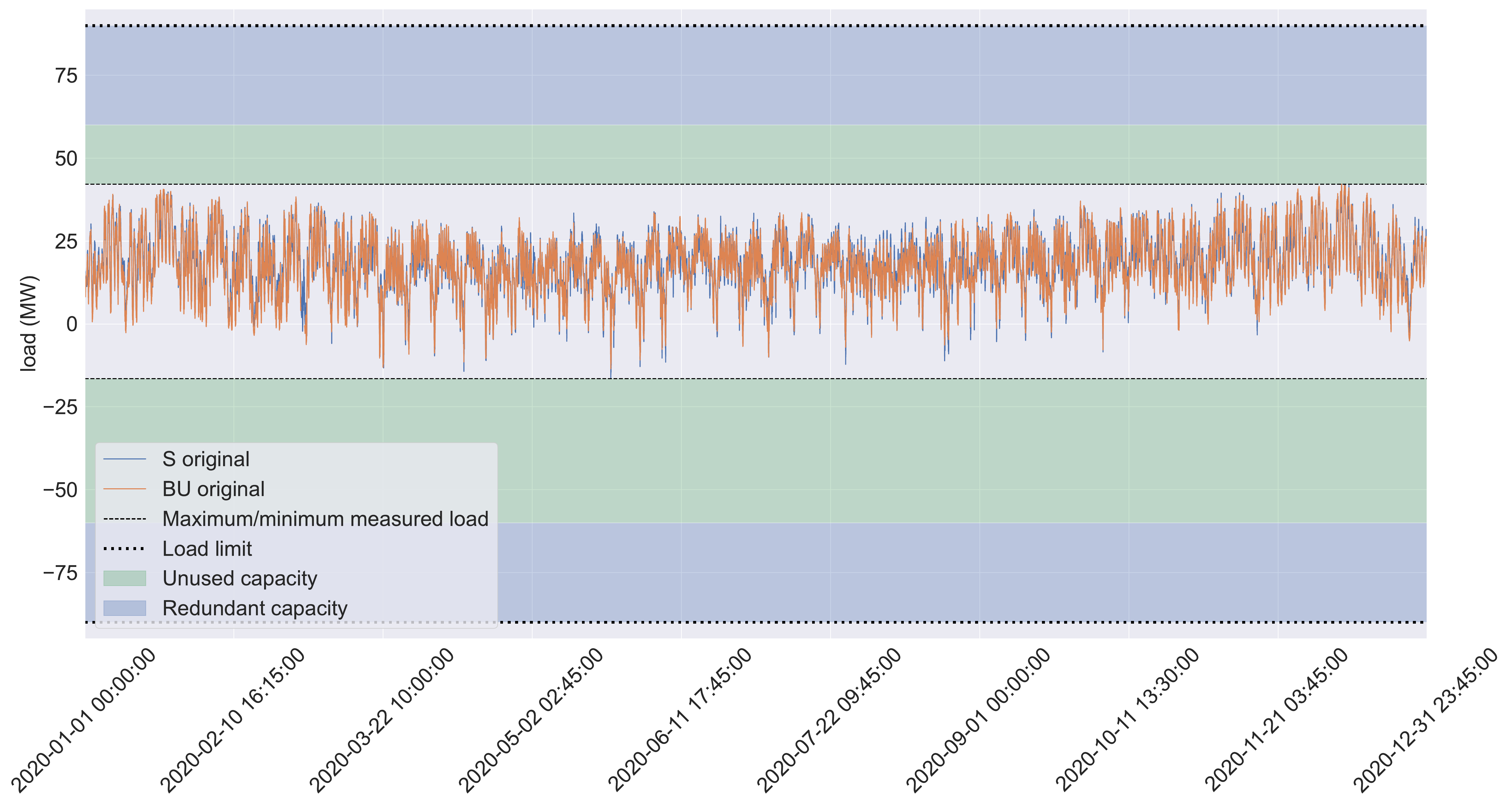}
\caption{A plot of the measured load ($S$) and the bottom-up load ($B$) as measured or estimated over the entire year for station 005. The S measurement is visualized in blue, and the bottom-up load is visualized in orange. The minimum and maximum load estimates are shown by the dashed lines. The load limit of the primary substation is shown by the dotted line. The green and blue areas indicate the unused and redundant capacity, these are fictitious and only shown for illustrative purposes.} 
\label{fig:measurement_example}
\end{figure}

\begin{figure}
\includegraphics[width=\textwidth]{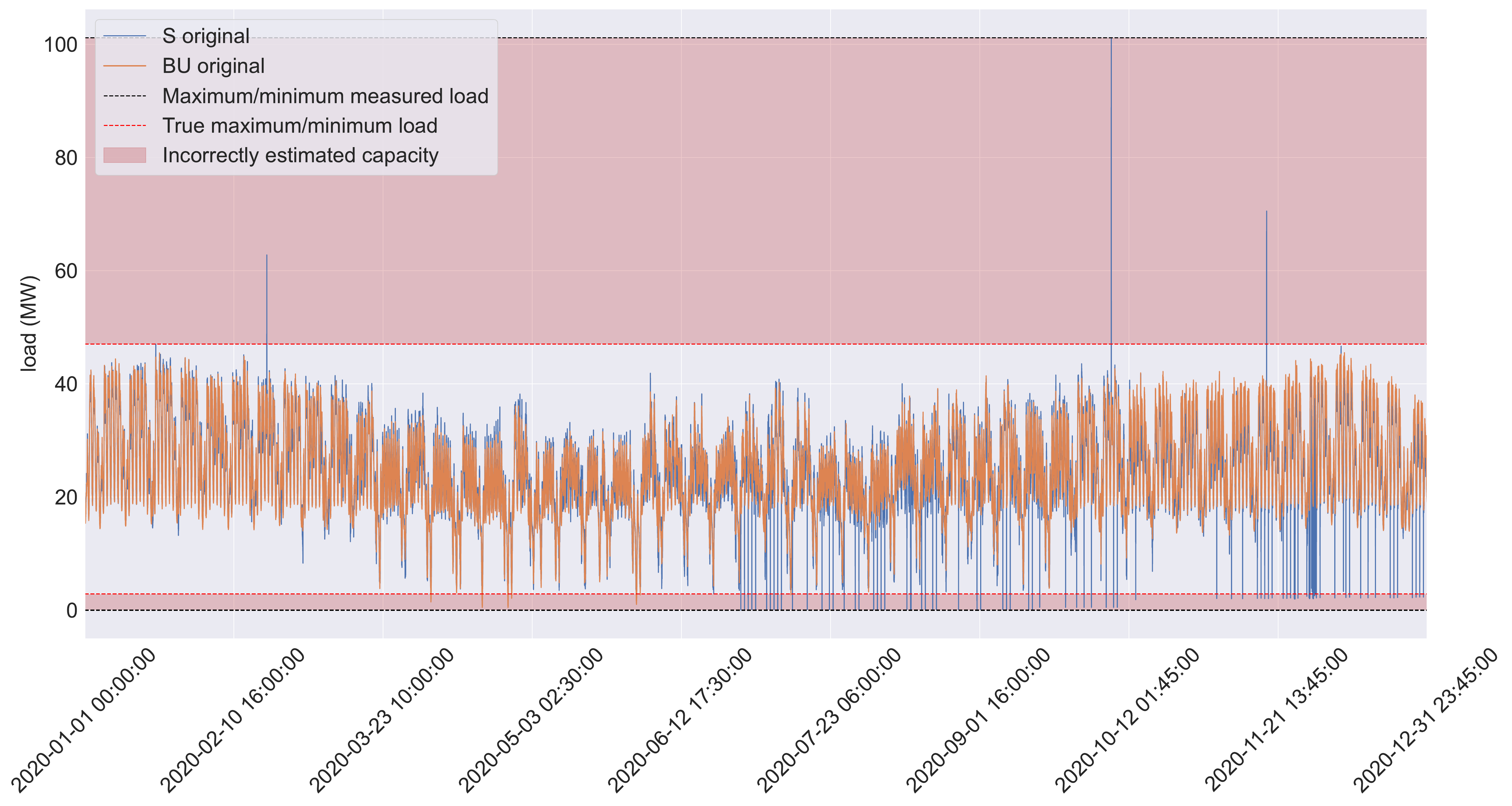}
\caption{A plot of the measured load ($S$) and the bottom-up load ($B$) as measured or estimated over the entire year for station 010. The S measurement is visualized in blue, and the bottom-up load is visualized in orange. The minimum and maximum load estimates are shown by the black dashed lines. The load limit of the primary substation is shown by the dotted line. The true minimum and maximum load limits are shown by the red dashed lines. The capacity that would be incorrectly included in the estimate is shown by the opaque red boxes.} 
\label{fig:anomaly_example}
\end{figure}

In order to discern between a primary substation connection that is net consuming or net producing, the load measurements $S$ are given a sign based on $P$. If a primary substation connection is consuming more than it is producing, the load measurement is assigned a positive sign. A negative sign therefore means that the primary substation is net producing.  However, not all primary substations are outfitted with measurement equipment to determine whether the primary substation is net producing or consuming: they just measure the absolute current $I$, thus the sign needs to be corrected later on. The bottom-up load, in contrast to the actual load, can always be measured in the negative due to being based on more recent, lower-level, measuring equipment. We will use this property in correcting the sign of the load measurement. Bottom-up load measurements are furthermore based on $P$ measurements and load estimates so they never have a missing sign.
A typical primary substation measurement time series, consisting out of  the load ($S$), the bottom-up load, and an illustrative example of the needed minimum and maximum capacity is shown in Figure \ref{fig:measurement_example}, where it should be noted that this measurement does not contain anomalies or switch events. In this figure we can additionally see the minimum and maximum loads ($S$) that are vitally important for the planning of grid expansion. On top of the maximum or minimum load, there should be enough redundant capacity in order to allow for rerouting in case of grid failures. All remaining capacity is unused, and quantification of this unused capacity is essential for determining whether the capacity of a station should be expanded, or whether there is room for additional customers. Should anomalies or switch events be present, we expect to see these minimum and maximum values be inflated or deflated, leading to inaccurate estimates of the unused capacity of a primary substation, thereby interfering with power grid expansion planning. An example of wrong capacity estimates leading to unused capacity can be found in Figure~\ref{fig:anomaly_example}.

To summarize, at each 15-minute interval $t$ for each station $i$ we measure the load $s_t^{i}$ and estimate the bottom-up load $b_t^{i}$. We do this for a full year of measurements, yielding for each station $i$ a load vector $\vect{s}^{i} \in S$, and a bottom-up load vector $\vect{b}^{i} \in B$, where $S$ and $B$ respectively now denote the full collections of load and bottom-up vectors as measured for each of the \nOSmeasurements measured stations.
In order to allow for evaluation of automatic segmentation and anomaly detection algorithms, each 15-minute measurement $y_t^{i}$ has additionally been labeled, leading to a detailed labeled segmentation of each time series for each station $i$, $\vect{y}^{i} \in Y$, which we will treat as our gold standard. Possible values for this label are: $0$ (no anomaly or switch event); $1$ (anomaly or switch event); $5$ (the labeler is uncertain whether this should be label $0$ or $1$). For simplicity, the dependence of vectors $\vect{s}$, $\vect{b}$ and $\vect{y}$ on index of the station $i$ will be omitted throughout the remainder of this paper, and will only be mentioned explicitly when necessary.

The time series of all primary substations show great variation, both within a primary substation, but also between primary substations. The average load for a primary substation can range from 100's to 10,000's of kilowatts. In addition, not all bottom-up loads are equally accurate for each primary substation. This variability leads to a challenging segmentation task.

\subsubsection{Event-length Categories}
\begin{figure}
\includegraphics[width=\textwidth]{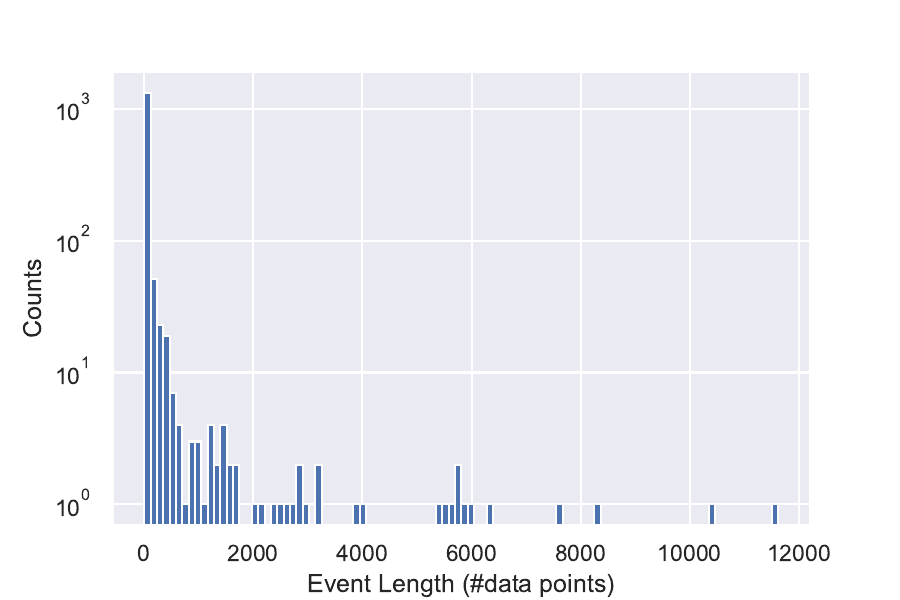}
\caption{Histogram of the length of the events and anomalies over all datasets. Note that the y-axis is log-scaled due to the frequency of short events. A year typically consists out of 35040 15-minute interval measurements.} 
\label{fig:event_length_distribution}
\end{figure}

We use a variety of measures to judge the quality of the time series segmentation, see Section~\ref{section:evalutation_metrics}. Generally, these measures are calculated over all individual time points. This can be done when the segments are of somewhat similar length. In this use case however, anomalies are very short events, while switch events might be very long. This disparity in event length is illustrated in Figure \ref{fig:event_length_distribution}. From this figure, we can clearly observe that long events are rare, but make up the majority of measured data labeled as $1$, while short events/anomalies are much more frequent, but only make up a small part of the label $1$ data. In order to alleviate this problem while optimizing and evaluating our methods, we divide the anomaly/switch events in 4 categories based on their length based on how many samples they consists of, and calculate all measures, see Section~\ref{section:evalutation_metrics} for each category.
The 4 categories of event lengths are defined as: 1 up to and including 24 samples, 25 up to including 288 samples, 289 up to and including 4032 samples, and 4033 samples and longer. From here on out we will refer to these, for sake of clarity, as their equivalents in time units: 15 minutes to 6 hours, 6 hours to 3 days, 3 to 42 days, and 42 days or longer. Roughly speaking the categories contain the following types of events: 15 minutes to 6 hours generally contains measurement errors, 6 hours to 3 days contains longer measurement errors and very short switch events that are easily resolved, 3 to 42 days contains short switch events, as well as more complex rerouting, and 42 days or longer contains switch events caused by for example complex rerouting, long-term maintenance, and grid expansion. It should be noted that this categorization is not set in stone, and may be chosen slightly differently depending on the prevalence of event types in other applications.

\subsection{Preprocessing}

In order to make each time series suitable for further analysis, several preprocessing steps have been conducted. First, missing $S$ measurements and corresponding bottom-up loads are removed from each time series. Missing data can occur when there is a communication error in the system. Communication errors are typically characterized by repeated measurements or bottom-up loads.  To correct for the discrepancies between the load measurement and the bottom-up load, we perform a linear regression to better match the two. The multiplication term corrects for multiplicative mismatches caused by over- or underestimating the amount of load of the customers and the grid losses that depend on the load (copper-losses), while the additive baseline correction corrects for constant grid losses, or iron losses, and, in the case of a current measurement, for the constant reactive power caused mainly by the capacitance of the cables. The linear regression is done on a subset of the time series, specifically by excluding everything outside user-defined quantiles $q_{\textrm{lower}}\%$ and $q_{\textrm{upper}}\%$. 
This is done to prevent that any anomalies or switch events steer the linear regression. Lastly, we perform a sign correction on $S$ measurements as some measurement equipment cannot measure load signs.
We list the preprocessing steps as pseudocode in Algorithm \ref{code:preprocessing} in the order in which they are applied to each individual station measurement, i.e., all $\vect{s} \in S$ and $\vect{b} \in B$. Note that $\vect{s}^i$ and $\vect{b}^i$ are vectors of equal length. The used functions ``bottomUpMissing'' and ``repeatedMeasurements'' are described in Algorithm~\ref{code:bottom_up} and Algorithm~\ref{code:repeated_measurements} respectively.

\begin{algorithm}
\caption{Preprocessing procedure}\label{code:preprocessing}
\begin{algorithmic}[1]
\State \textbf{Input:} Measurement $\vect{s}$ and bottom-up load $\vect{b}$
\State \textbf{Hyperparameters:} The maximum number of repeated measurements $r$,
and the quantile boundaries used for scaling before the linear fit $q_{\textrm{lower}}\%$, $q_{\textrm{upper}}\%$
\State \textbf{Output:} Difference vector $\vect{\delta}$
\\

\For{$i \gets 1$ to $n$} \Comment{Remove missing and repeated measurements}
    \If{$\textrm{bottomUpMissing}(b_i) ~\textbf{or}~ \textrm{repeatedMeasurements}(s, i, r)$}
        \State Remove $s_i$ from $\vect{s}$ and $b_i$ from $\vect{b}$ \Comment{See Algorithm \ref{code:bottom_up} and  \ref{code:repeated_measurements}.}
    \EndIf
\EndFor
    
\State  $\vect{\delta}_{\textrm{temp}} \gets \vect{s} - \vect{b}$ \Comment{Calculate the temporary difference vector}

\State $q_{\textrm{min}} \gets \textrm{quantile}(\vect{s}, q_{\textrm{lower}}\%)$
\State $q_{\textrm{max}} \gets \textrm{quantile}(\vect{s}, q_{\textrm{upper}}\%)$

\State $\vect{s}_{\textrm{filtered}} \gets s_i \in \vect{s}$ where $s_i > q_{\textrm{min}}$ and $s_i < q_{\textrm{max}}$ 
\State $\vect{b}_{\textrm{filtered}} \gets$ corresponding elements $\vect{b} \in B$

\State $\vect{s}_{\textrm{filtered}} = m \cdot \vect{b}_{\textrm{filtered}} + c$ \Comment{Fit linear model to find slope $m$ and offset $c$}

\State $\vect{b}_{\textrm{scaled}} \gets m \cdot \vect{b} + c$ \Comment{Rescale the bottom-up to match $S$}

\\ \Comment{Correct the sign of $\vect{s}$ if the minimum of $s$ is positive while the minimum of $b$ is negative.}
\If{$\textrm{min}(\vect{s}) \geq 0$ \textbf{and} $\textrm{min}(\vect{b}_{\textrm{scaled}}) < 0$}
    \State $\vect{s}_{\textrm{signed}} \gets \vect{s} \cdot \textrm{sign}(\vect{b}_{\textrm{scaled}})$
\Else
    \State $\vect{s}_{\textrm{signed}} \gets \vect{s}$
\EndIf

\State $\vect{\delta} \gets \vect{s}_{\textrm{signed}} - \vect{b}_{\textrm{scaled}}$ \Comment{Calculate the difference vector}

\end{algorithmic}
\end{algorithm}

\begin{algorithm}
\caption{bottomUpMissing}\label{code:bottom_up}
\begin{algorithmic}[1]
\State \textbf{Input:} A single bottom-up measurement $b_i$
\State \textbf{Output:} Boolean $\beta$ indicating whether the bottom-up load measurement $b_i$ is missing
\\
\State \[
 \beta \gets
 \begin{cases}
     \textrm{True}, & \text{if } b_i = \text{NaN}\\
     \textrm{False}, & \text{otherwise}
 \end{cases}
\]

\end{algorithmic}
\end{algorithm}

\begin{algorithm}
\caption{repeatedMeasurements}\label{code:repeated_measurements}
\begin{algorithmic}[1]
\State \textbf{Input:} Measurement $\vect{s}$ and index $i$
\State \textbf{Hyperparameters:} The maximum number of repeated measurements $r$
\State \textbf{Output:} Boolean $\beta$ indicating whether the maximum allowed number of repeated measurements and adjacent to $s_i$ is exceeded 
\\
\State $\beta \gets \textrm{False}$

\State $c \gets 0$ \Comment{Repeated measurement counter}
\State $\textit{max\_c} \gets 0$
\For{$j \gets \max(i-r, 0)$ to $\min(i+r, n)$} \Comment{$n$ is the length of the vector $\vect{s}$}
    \If{$s_j = s_i$}

        \State $\textit{max\_c} \gets \max(c, \textit{max\_c})$
    \Else
         \State $c \gets 0$
    \EndIf
\EndFor

\If{$\textit{max\_c} \geq r$}
    \State $\beta \gets \textrm{True}$
\EndIf

\end{algorithmic}
\end{algorithm}

As one can note, this procedure has several user-defined hyperparameters, which can be optimized. These are specifically the range of the quantiles, $q_{\textrm{lower}}\%$ and $q_{\textrm{upper}}\%$, used in the filtering procedure, as well as the number of $\vect{s}$ measurements, $r$, that have to be identical in order to be classified as missing due to a communication error. Based on manual observations of the preprocessing procedure, we have selected $q_{\textrm{lower}}\%$ and $q_{\textrm{upper}}\%$ to be $10\%$ and $90\%$ respectively, and $r$ to be $5$.

The resulting difference vector $\vect{\delta}$ is calculated for each primary substation, yielding the set of difference vectors $\Delta$, and then used for further analysis. $\vect{\delta}$ now represents the error between the actual measurement and the bottom-up load. Effectively we now have a feature vector where the variation that can be explained from the bottom-up load has been removed from the $S$ measurement. This difference vector is used as input for the various segmentation and anomaly detection algorithms we apply, all listed in section~\ref{section:algorithms}.

\subsection{Algorithms and Optimization}
\label{section:algorithms}
In this research, we compare several methods for detecting anomalies and switch events. Each of these takes the difference vector between the measurement and the bottom-up load $\vect{\delta}^i \in \Delta$ as input data. The output of each base method is a vector of unbounded scores for each primary substation $\vect{z}^i$, thus yielding a set of score vectors $Z$. These scores are subsequently converted to predicted binary label vectors for each primary substation $\vect{\tilde{y}}^i$, yielding a set of predicted label vectors $\tilde{Y}$. This is done by thresholding the scores, on which more details can be found in section \ref{section:threshold_optimization}.

We specifically make use of unsupervised anomaly detection algorithms. Unsupervised algorithms do not learn from labeled data, but only consider the measurements. We have annotated a fairly large number of \nOSmeasurements yearly primary substation measurements, but recognize that anomalies and switch events are very rare and heterogeneous events. Because of this, we applied unsupervised methods of detection. We then assume that some higher level hyperparameters of the unsupervised procedure, specifically the thresholds used for acquiring labels, will generalize towards unobserved types of anomalies and switch events, we will evaluate this assumption on the test set.

We have applied 3 base detectors, specifically statistical process control, isolation forest, and binary segmentation, and describe them in section~\ref{section:base_detectors}. We compare these detectors separately, but also ensemble them by combining them in several ways. We compare a naive ensemble method, a distinct optimization criterion ensemble, and a sequential ensemble, which we describe in section~\ref{section:ensembles}.

\subsubsection{Base Detectors}
\label{section:base_detectors}
\paragraph{Statistical process control}
Under the assumption that the data is stationary we can use classical statistical process control, or SPC, methods~\citep{oakland2007statistical} in order to detect anomalies and switch events. In traditional SPC one assumes a process which is stationary within a chosen time frame or segment. Then, the user defines certain lower and upper control limits, typically 2 or 3 standard deviations away from the mean of the segment. In our case, the control limits will be optimized as either symmetric or asymmetric thresholds, rather than using the traditional statistical approach. Additionally, due to the known presence of anomalies in the data, we also look at the distance to the median, rather than to the mean. We will explicitly refer to the optimized control limit as ``threshold'' from here on out. When a time point falls outside of these control limits, it is flagged as out-of-control, in this case meaning anomalous. How SPC is applied to each difference vector $\vect{\delta} \in \Delta$ resulting from preprocessing is described in Algorithm~\ref{code:SPC}

\begin{algorithm}
\caption{Statistical process control}
\label{code:SPC}
\begin{algorithmic}[1]
\State \textbf{Input:} Difference vector $\vect{\delta}$
\State \textbf{Hyperparameters:} Quantile boundaries for scaling $q_{\textrm{lower}}\%$, $q_{\textrm{upper}}\%$ 
\State \textbf{Output}: Score vector $\vect{z}$ 
\\
\State $m \gets \textrm{median}(\vect{\delta})$ \Comment{Calculate median}
\State $d \gets \textrm{interquantileDistance}(\vect{\delta}, q_{\textrm{lower}}\%$, $q_{\textrm{upper}}\%)$ \Comment{Calculate interquantile distance}
\State $\vect{z} \gets  (\vect{\delta}-m)/d$
\end{algorithmic}
\end{algorithm}
It should be noted that the relevant hyperparameters here are the quantiles that are chosen for the interquantile range. These can be distinct from the quantiles hyperparameters used in the preprocessing procedure.

\paragraph{Isolation forest}
One of the most commonly used machine learning methods for anomaly detection is the isolation forest, or IF~\citep{liu2008isolation}. The isolation forest is known as one of the best state-of-the-art anomaly detectors on real-valued static data~\citep{Bouman2024UnsupervisedADComparison}. An isolation forest on one-dimensional data effectively produces a density estimate by randomly splitting subsets of the data. We consider two distinct ways of applying isolation forests on the data: one where we apply a single forest per station difference vector, see Algorithm~\ref{code:forest_per_station}, and one where we scale and concatenate all difference vectors for training the isolation forest and apply that isolation forest on each difference vector, see Algorithm~\ref{code:single_forest}. Note that in these algorithms, we use several high-level functions. ``fitPredictIsolationForest'' fits an isolation forest with $n_{\textrm{estimators}}$ trees on the input $\vect{\delta}$ and returns the anomaly scores $\vect{\tilde{z}}$ on the same input. ``fitIsolationForest'' fits an isolation forest with $n_{\textrm{estimators}}$ trees on the input $\vect{\delta}$ and returns the fitted model $\gamma$. ``predictIsolationForest'' returns the anomaly scores $\vect{\tilde{z}}^i$ calculated over $\vect{\delta}_{\textrm{scaled}}^i$ given an already fitted isolation forest $\gamma$. Each of these functions is implemented as part of the scikit-learn library~\citep{scikit-learn}.

\begin{algorithm}
\caption{Isolation forest per station}
\label{code:forest_per_station}
\begin{algorithmic}[1]
\State \textbf{Input:} Difference vector $\vect{\delta}$
\State \textbf{Hyperparameters:} The number of trees $n_{\textrm{estimators}}$
\State \textbf{Output}: Score vector $\vect{z}$
\\
\State $\vect{\tilde{z}} \gets  \textrm{fitPredictIsolationForest}(\vect{\delta}, n_{\textrm{estimators}})$
\State $\vect{z} \gets -\vect{\tilde{z}} + 1$ \Comment{Rescale so higher score means more anomalous}
\end{algorithmic}
\end{algorithm}

\begin{algorithm}
\caption{Single isolation forest over all stations}
\label{code:single_forest}
\begin{algorithmic}[1]
\State \textbf{Input:} Difference vectors $\vect{\delta}^i \in \Delta$ for each station
\State \textbf{Hyperparameters:} The number of trees $n_{\textrm{estimators}}$, quantile boundaries for scaling $q_{\textrm{lower}}\%$, $q_{\textrm{upper}}\%$  
\State \textbf{Output:} Score vectors $\vect{z}^i \in Z$ for each station 
\\

\For{$\vect{\delta}^i \in \Delta$}
    \State $m \gets \textrm{median}(\vect{\delta}^i)$ 
    \State $d \gets \textrm{interquantileDistance}(\vect{\delta}^i, q_{\textrm{lower}}\%$, $q_{\textrm{upper}}\%)$ 
    \State $\vect{\delta}_{\textrm{scaled}}^i \gets (\vect{\delta}^i-m)/d$
    
\EndFor
\State $\Delta_{\textrm{scaled}} \gets \{\vect{\delta}_{\textrm{scaled}}^i, ..., \vect{\delta}_{\textrm{scaled}}^n\}$

\State $\gamma \gets \textrm{fitIsolationForest}(\Delta_{\textrm{scaled}}, n_{\textrm{estimators}})$

\For{$\vect{\delta}_{\textrm{scaled}}^i \in \Delta_{\textrm{scaled}}$}
    \State $\vect{\tilde{z}}^i \gets  \textrm{predictIsolationForest}(\gamma, \vect{\delta}_{\textrm{scaled}}^i)$
    \State $\vect{z}^i \gets -\vect{\tilde{z}}^i + 1$ \Comment{Rescale so higher score means more anomalous}
\EndFor

\end{algorithmic}
\end{algorithm}
In either procedure, we need to set the hyperparameters of the isolation forest. In the case where a single isolation forest is applied, we also consider the quantile boundaries for scaling. Furthermore, the choice between fitting a single forest or a forest per station is treated as a hyperparameter. We perform rescaling of the scores so that the algorithms deems the most anomalous samples to have the highest scores. These definitions differ between different algorithms in literature, but we chose this definition to be analogous to SPC, allowing for a similar threshold optimization strategy.

\paragraph{Binary segmentation}
Binary segmentation is a change point detection algorithm able to find multiple change points in a given time series~\citep{scott1974cluster,bai1997estimating}. We use binary segmentation as a state-of-the-art change point algorithm because it is found to be one of the best performing algorithms on real-world univariate time series data~\citep{van2020evaluation, clasp2021, clasp2023}. It finds change points by recursively partitioning the time series into two parts, forming a binary tree. A split occurs at the optimal break point. This break point is found by first calculating the cost $c_{\textrm{total}}$ of the entire segment using a chosen cost function. Then the costs of the two subsegments, $c_{\textrm{left}}$ and $c_{\textrm{right}}$, are calculated for each possible break point using the same cost function. The optimal break point is then found by selecting the one for which the gain, $g = c_{\textrm{total}} - c_{\textrm{left}} - c_{\textrm{right}}$ is maximized. This procedure is repeated until the gain for a split is below a user-defined threshold called the penalty $p$. In our experiments, we consider two penalties, a linear and a L1 penalty~\citep{van2020evaluation}, the scaling of which depends on a user defined $\beta$ parameter.

We use binary segmentation in order to generate scores by performing the following procedure explained in Algorithm~\ref{code:binary_segmentation}. The ``findReferenceValue'' function is described in more detail in Algorithm~\ref{code:find_reference_value}. The ``findBreakpointsBinarySegmentation'' is a call to the high-level ``ruptures'' Python library~\citep{truong2018ruptures} which takes the scaled vector $\vect{\tilde{z}}$ and segments it according to the well-known binary segmentation algorithm with hyperparameters $\beta, C, l$, and $j$.

\begin{algorithm}
\caption{Binary segmentation}
\label{code:binary_segmentation}
\begin{algorithmic}[1]
\State \textbf{Input:} Difference vector $\vect{\delta}$
\State \textbf{Hyperparameters:} Cost function $C$, cost function weight $\beta$, minimum segment size $l$, the jump size $j$, quantile boundaries for scaling $q_{\textrm{lower}}\%$, $q_{\textrm{upper}}\%$, and the reference point strategy $\textit{reference\_point}$ 
\State \textbf{Output}: Score vector $\vect{z}$
\\

\State $m \gets \textrm{median}(\vect{\delta})$ 
\State $d \gets \textrm{interquantileDistance}(\vect{\delta}, q_{\textrm{lower}}\%$, $q_{\textrm{upper}}\%)$
\State $\vect{\tilde{z}} \gets  (\vect{\delta}-m)/d$
\\
\State $\vect{b} \gets  \textrm{findBreakpointsBinarySegmentation}(\vect{\tilde{z}}, \beta, C, l, j)$

\State $r \gets \textrm{findReferenceValue}(\vect{\tilde{z}}, \vect{b}, \textit{reference\_point})$
\State $b_{\textrm{begin}} = 1$ \Comment{indexing begins at 1}
\For{$b_{\textrm{end}} \in \vect{b}$}

    \State $\vect{t} \gets (\vect{\tilde{z}}_i)_{b_{\textrm{begin}} \le i \le b_{\textrm{end}}}$ \Comment{Get segment}
    \State $(\vect{z}_i)_{b_{\textrm{begin}} \le i \le b_{\textrm{end}}} \gets \textrm{mean}(\vect{t}) - r$ \Comment{Get difference between segment and reference value}
    \State $b_{\textrm{begin}} = b_{\textrm{end}}$ 
\EndFor
\end{algorithmic}
\end{algorithm}

\begin{algorithm}
\caption{findReferenceValue}\label{code:find_reference_value}
\begin{algorithmic}[1]
\State \textbf{Input:} Scaled difference vector $\vect{\tilde{z}}$, and breakpoints resulting from binary segmentation $\vect{b}$
\State \textbf{Hyperparameters:} Reference point strategy $\textit{reference\_point}$
\State \textbf{Output:} Reference point value $r$
\\

\If{$\textit{reference\_point} = \textit{``mean''}$}
    \State $r = \textrm{mean}(\vect{\tilde{z}})$
\ElsIf{$\textit{reference\_point} = \textit{``median''}$}
    \State $r = \textrm{median}(\vect{\tilde{z}})$
\ElsIf{$\textit{reference\_point} = \textit{``longest\_mean''}$}
    \State $s_{\max} = 0$
    \State $b_{\textrm{begin}} = 1$ \Comment{indexing begins at 1}
    \For{$b_{\textrm{end}} \in \vect{b}$}
        \State $\vect{t} \gets (\vect{\tilde{z}}_i)_{b_{\textrm{begin}} \le i \le b_{\textrm{end}}}$ \Comment{Get segment}
        \State $s_i \gets b_{\textrm{end}} -  b_{\textrm{begin}}$ \Comment{Get segment size}
        \If{$s_i > s_{\max}$}
            \State $s_{\max} = s_i$
            \State $r = \textrm{mean}(\vect{t})$
        \EndIf
        \State $b_{\textrm{begin}} = b_{\textrm{end}}$ 
    \EndFor
\ElsIf{$\textit{reference\_point} = \textit{``longest\_median''}$}
    \State $s_{\max} = 0$
    \State $b_{\textrm{begin}} = 1$ \Comment{indexing begins at 1}
    \For{$b_{\textrm{end}} \in \vect{b}$}
        \State $\vect{t} \gets (\vect{\tilde{z}}_i)_{b_{\textrm{begin}} \le i \le b_{\textrm{end}}}$ \Comment{Get segment}
        \State $s_i \gets b_{\textrm{end}} -  b_{\textrm{begin}}$ \Comment{Get segment size}
        \If{$s_i > s_{\max}$}
            \State $s_{\max} = s_i$
            \State $r = \textrm{median}(\vect{t})$
        \EndIf
        \State $b_{\textrm{begin}} = b_{\textrm{end}}$ 
    \EndFor
\EndIf
\end{algorithmic}
\end{algorithm}

This procedure has a large number of optimizable hyperparameters resulting from the binary segmentation algorithm, the strategy for determining the reference point, and with which quantiles scaling should be applied. We have compared 4 different strategies for determining the reference point: 1) comparing to the mean of the entire scaled station difference vector ($\textit{``mean''}$), 2) comparing to the median of the entire scaled station difference vector($\textit{``median''}$), 3) comparing to the mean of the longest segment ($\textit{``longest\_mean''}$), and 4) comparing to the median of the longest segment ($\textit{``longest\_median''}$).

\subsubsection{Ensembles}
\label{section:ensembles}
The different base detection algorithms have various strengths and weaknesses. Specifically, binary segmentation is good at detecting long events, while SPC and IF are good at detecting shorter events. Binary segmentation more easily detects long events as it can compare the distance of an entire segment to the distance of a different segment, thus being more sensitive to small changes over a long time period. SPC and IF consider each timepoint individually, thus only detecting large changes without considering the time component. In order to leverage the strengths of multiple complementary methods, we employ different ensembling techniques to combine base detection algorithms. We specifically compare naive ensembling through combining predictions directly, combining algorithms optimized on detecting events of different lengths, and sequential ensembles, where we apply binary segmentation and apply SPC or IF only on the ``normal" segments. A more detailed description of each ensembling strategy can be found below. For all ensembles, we look at a combination of a change point detector, binary segmentation, and an anomaly detector, which is either SPC or IF. As both SPC and IF are designed to find singular anomalies, we see them as complementary to binary segmentation, and will specifically compare SPC to IF.

\paragraph{Naive Ensembles}
\label{section:naive_ensembles}
The simplest way of combining base detection algorithm predictions is by what we call naive ensembling. In naive ensembling, we simply take the predictions of each base detection algorithm, and combine them using an OR operation, meaning when either algorithm predicts an anomaly or switch event, so does the ensembled prediction.

\paragraph{Different Optimization Criterion Ensembles}
\label{section:different_optimization_ensembles}
As noted earlier, it is known that binary segmentation is good at detecting long events, while SPC and IF are good at detecting shorter events. By simply combining predictions as in naive ensembling, the individual strengths of the sub-models are not leveraged to their fullest extent. We can instead optimize both detection models on different criteria. In this case, we have optimized binary segmentation to detect longer events in the ``3 to 42 days", and ``42 days and longer" length categories, whereas we optimize either SPC or IF on the ``15 minutes to 6 hours", and ``6 hours to 3 days" length categories. Then, like in naive ensembling, we combine the predictions of both separately optimized algorithms using an OR operation. Due to the specific optimization we expect this method to find fewer false positives and have a higher recall. For clarity's sake, we will abbreviate this type of ensemble as DOC, Different Optimization Criterion, ensembles.

\paragraph{Sequential ensembles}
\label{section:sequential_ensembles}
Sequential ensembles follow the same idea of using differently optimized detectors. In a sequential ensemble we first apply binary segmentation optimized on the ``3 to 42 days", and ``42 days and longer" length categories. Then, all segments that were not classified as switch events are passed to a second detection algorithms, either SPC or IF, to detect shorter events and anomalies. The second method is again optimized for detecting the shorter event categories ``15 minutes to 6 hours", and ``6 hours to 3 days". Due to the further specificity of this method, we expect to find fewer false positives, and to have a higher recall, especially on the shorter events which are no longer being masked by long switch events.

\subsection{Evaluation and Optimization}
\label{section:evaluation_optimization}

\subsubsection{Evaluation Metrics}
\label{section:evalutation_metrics}
Typically, time series segmentation is evaluated using precision, recall, the ROC/AUC, and the F$\beta$ score~\citep{clasp2021,clasp2023}. In this research, we focus on a weighted approach of the precision, recall, and the F$\beta$ score. Rather than calculating them directly on a sample-to-sample basis, we calculate them for each of the 4 defined length categories. We do this so we can accurately detect events across the entire spectrum of lengths, as the number of measurements of events in each category increases greatly for increasing event lengths. For each of the length categories, we calculate the precision, recall, and the F$1.5$ score. We use $\beta = 1.5$ to give a higher importance to the recall term, as the potential impact of a false negative is higher than that of a false positive in power grid expansion planning. When we calculate a score for a length category, we do not include timestamps where either the assigned label is 5 (``uncertain"), or where the assigned label is 1 for any of the other categories. In order to get an estimate of the overall performance of the model, we average these measures over the four categories. The metric used to optimize any (hyper)parameter is the average F$1.5$ score over all 4 length categories. We explicitly choose to study the precision, recall, and the F$\beta$ scores. In the practical use case we describe, the predictions are always binary, even though they are based on real-valued scores. As we explicitly threshold the scores to labels in order to filter our data, the precision, recall, and the F$\beta$ score most closely illustrate the performance trade-off that is being made by thresholding. While we do not explicitly study the ranking of the anomalies and switch events, as is commonly done by the ROC/AUC metric, we do provide an additional plot of the performance in terms of the ROC/AUC in \ref{appendix:AUROC}.

\subsubsection{Validation}
In order to optimize thresholds, hyperparameters and to compare models, we split our original dataset consisting out of \nOSmeasurements stations into 3 equal parts of 60 stations each, creating a training, validation, and test dataset. This splitting procedure was done in a stratified manner such that the three splits have a more or less equal distribution of event lengths. The stations were divided by aiming to have an equal number of events for each category present in each dataset. The computational complexity of evaluating all possible combinations, like is done in available cross-validation software, is too high. We have therefore used a greedy approach to produce a stratified split of the data, where stations were divided starting with the longest, and least frequent, event length category, and ending with the shortest and most frequent event length category. All stations with no events were subsequently divided amongst the datasets in order to create 3 equally large datasets. The number of events, as well as the label ``1" counts, can be found in Table~\ref{table:event_length_stats}.

\begin{table*}
\caption{The distribution of event lengths over the train, test, and validation splits. The event count indicates how many events within that category are in a dataset. The label ``1" count indicates how many separate time points belong to the anomaly/switch event class per dataset. }

\label{table:event_length_stats}
\centering

\resizebox{\columnwidth}{!}{%
\begin{tabular}{llrrrr}
\toprule
 &  & 15m-6h & 6h-3d & 3d-42d & 42d and longer \\
 & Dataset &  &  &  &  \\
\midrule
\multirow[t]{4}{*}{Event count} & Train & 338 & 136 & 24 & 4 \\
 & Validation & 203 & 173 & 25 & 4 \\
 & Test & 444 & 99 & 23 & 4 \\
 & All & 985 & 408 & 72 & 12 \\
\cline{1-6}
\multirow[t]{4}{*}{Label $1$ count} & Train & 1506 & 6290 & 28386 & 28212 \\
 & Validation & 1971 & 13974 & 25075 & 30904 \\
 & Test & 2262 & 6790 & 27389 & 24992 \\
 & All & 5739 & 27054 & 80850 & 84108 \\
\cline{1-6}
\bottomrule
\end{tabular}

}
\end{table*}

From this table we can observe that the splitting of events in the longer categories is fairly balanced, whereas there is some imbalance in the shorter event categories. Even though the data is split in a stratified manner, such an imbalance can occur when single stations have a lot of events. This imbalance is most notable in the validation set, which has fewer ``15 minutes to 6 hours" events, but more ``6 hours to 3 days" events. Because these two categories are optimized on together, even within ensembles, we argue that this imbalance does not affect our conclusions.

After splitting, the train set is used to optimize the thresholds and train the isolation forest or just optimize the thresholds. The validation set is used to select the best performing hyperparameters for each method. Lastly, the test set is used to compare methods using only the best performing hyperparameters as evaluated on the validation set.

To get an estimate of the reliability of each method, we perform bootstrapping~\citep{efron1994introduction}. We do so by resampling the test set stations with replacement 10,000 times. In this way, we acquire both a bootstrapped mean and standard deviation for the precision, recall, and F$1.5$ metrics. 

\subsubsection{Threshold Optimization}
\label{section:threshold_optimization}
All methods we have applied produce scores $\vect{z} \in Z$. These can either be positives scores, where a higher value indicates a higher likelihood for a sample to be an anomaly or switch event according to the model, or scores centered around 0, where a greater distance to 0 indicates a higher likelihood. Of the presented methods, isolation forest has a purely positive score, while the statistical process control and binary segmentation scores are centered around 0. These scores are thresholded to yield a label prediction vector $\vect{\tilde{y}} \in \tilde{Y}$. This is done by applying a one-sided or symmetrical approach, $\vect{\tilde{y}} = \textrm{thresholdScores}(\textrm{abs}(\vect{z}), \theta^{\textrm{symmetrical}})$, or a two-sided or asymmetrical approach, $\vect{\tilde{y}} = \textrm{thresholdScores}(\vect{z}, \theta^{\textrm{lower}}, \theta^{\textrm{upper}})$. The symmetrical approach can be used on both positive and zero-centered scores, whereas the asymmetrical approach can only be used on the zero-centered scores. Pseudocode for the one-sided approach can be found in Algorithm~\ref{code:one_sided}, and for the two-sided approach in Algorithm~\ref{code:two_sided}.

\begin{algorithm}
\caption{thresholdScores (one-sided)}\label{code:one_sided}
\begin{algorithmic}[1]
\State \textbf{Input:} A score vector $\vect{z}$, where a higher $z_i$ indicates a higher likelihood of being an anomaly/switch event
\State \textbf{Hyperparameters:} A threshold $\theta^{\textrm{symmetrical}}$
\State \textbf{Output:} Predicted label vector $\vect{\tilde{y}}$
\For{$z_i \in \vect{z}$}
    \State \[
     \tilde{y}_i \gets
     \begin{cases}
         \textrm{1}, & \text{if } z_i \geq \theta^{\textrm{symmetrical}}\\
         \textrm{0}, & \text{otherwise}
     \end{cases}
    \]
\EndFor
\end{algorithmic}
\end{algorithm}

\begin{algorithm}
\caption{thresholdScores (two-sided)}\label{code:two_sided}
\begin{algorithmic}[1]
\State \textbf{Input:} A score vector $\vect{z}$, where a high or low $z_i$ indicates a higher likelihood of being an anomaly/switch event
\State \textbf{Hyperparameters:} A lower threshold $\theta^{\textrm{lower}}$, and upper threshold $\theta^{\textrm{upper}}$
\State \textbf{Output:} Predicted label vector $\vect{\tilde{y}}$
\For{$z_i \in \vect{z}$}
    \State \[
     \tilde{y}_i \gets
     \begin{cases}
         \textrm{1}, & \text{if } z_i \geq \theta^{\textrm{upper}}\\
         \textrm{1}, & \text{else if } z_i < \theta^{\textrm{lower}}\\
         \textrm{0}, & \text{otherwise}
     \end{cases}
    \]
\EndFor
\end{algorithmic}
\end{algorithm}

\begin{figure}
\includegraphics[width=\textwidth]{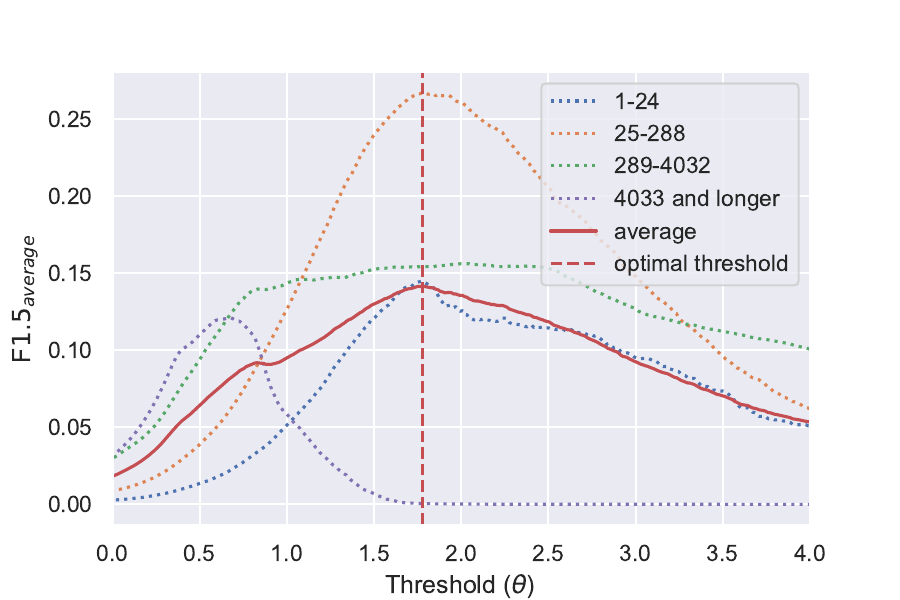}
\caption{Plot of the one-sided threshold optimization procedure. The F$1.5$ score, on the y-axis, as a function of the threshold, on the x-axis, is shown for all four distinct segment length categories, as well as their average. The red vertical line indicates the selected threshold which maximizes the F$1.5$ score on the average.} 
\label{fig:threshold_optimization}
\end{figure}

The thresholds are optimized by selecting those thresholds for which the average of the F$1.5$ over all 4 segment length categories is highest. This can be done efficiently by calculating the F$1.5$ score for each possible threshold value for all distinct segment length categories, and then averaging these profiles. An illustration of this selection procedure for the symmetrical approach can be found in Figure \ref{fig:threshold_optimization}. We can formalize this optimization as finding that threshold $\theta$, or those thresholds $\theta^{\textrm{lower}}, \theta^{\textrm{upper}}$ that maximize the average F$1.5$ score 
\[\theta_{\textrm{optimal}} = \argmax_{\substack{\theta}} \textrm{F}1.5_{\textrm{average}}(Y, \textrm{thresholdScores}(\textrm{abs}(Z), \theta))\]
in the symmetrical case, or 
\[\theta_{\textrm{optimal}}^{\textrm{lower}}, \theta_{\textrm{optimal}}^{\textrm{upper}} = \argmax_{\substack{\theta^{\textrm{lower}}, \theta^{\textrm{upper}}}} \textrm{F}1.5_{\textrm{average}}(Y, \textrm{thresholdScores}(Z, \theta^{\textrm{lower}}, \theta^{\textrm{upper}}))\]
in the two-sided case.

Whether to use one- or two-sided optimization is treated as a hyperparameter in the evaluation and selection process. 

\subsection{Implementation and Reproducibility}
Our analysis has been done in the Python programming language, specifically version 3.10.0. Many of our calculations rely on NumPy~\citep{harris2020array} and Pandas~\citep{the_pandas_development_team_2023_10107975}. We furthermore make use of the Ruptures\citep{truong2018ruptures} package for binary segmentation, and use the Scikit-learn~\citep{scikit-learn} package for scaling procedures, as well as applying the isolation forest. Visualizations were made using the Seaborn~\citep{Waskom2021} and Matplotlib~\citep{Hunter:2007} packages.
In order to reproduce all our experiments, we have provided access to a public GitHub repository\footnote{The Git repository can be found at: \href{https://github.com/RoelBouman/StormPhase2}{https://github.com/RoelBouman/StormPhase2}}. This reproduces the data splitting procedure, all experiments, including hyperparameter optimization, as well as producing all figures and tables in this paper. All data is provided through Alliander\footnote{The data repository can be found at \href{Alliander's open data repository.}{https://www.liander.nl/over-ons/open-data} (At time of review, this data is not yet publicly available)}.
An overview of all evaluated, and optimal, hyperparameters can be found in \ref{appendix:hyperparameters}.

\section{Results}

\begin{figure}

\includegraphics[width=\linewidth]{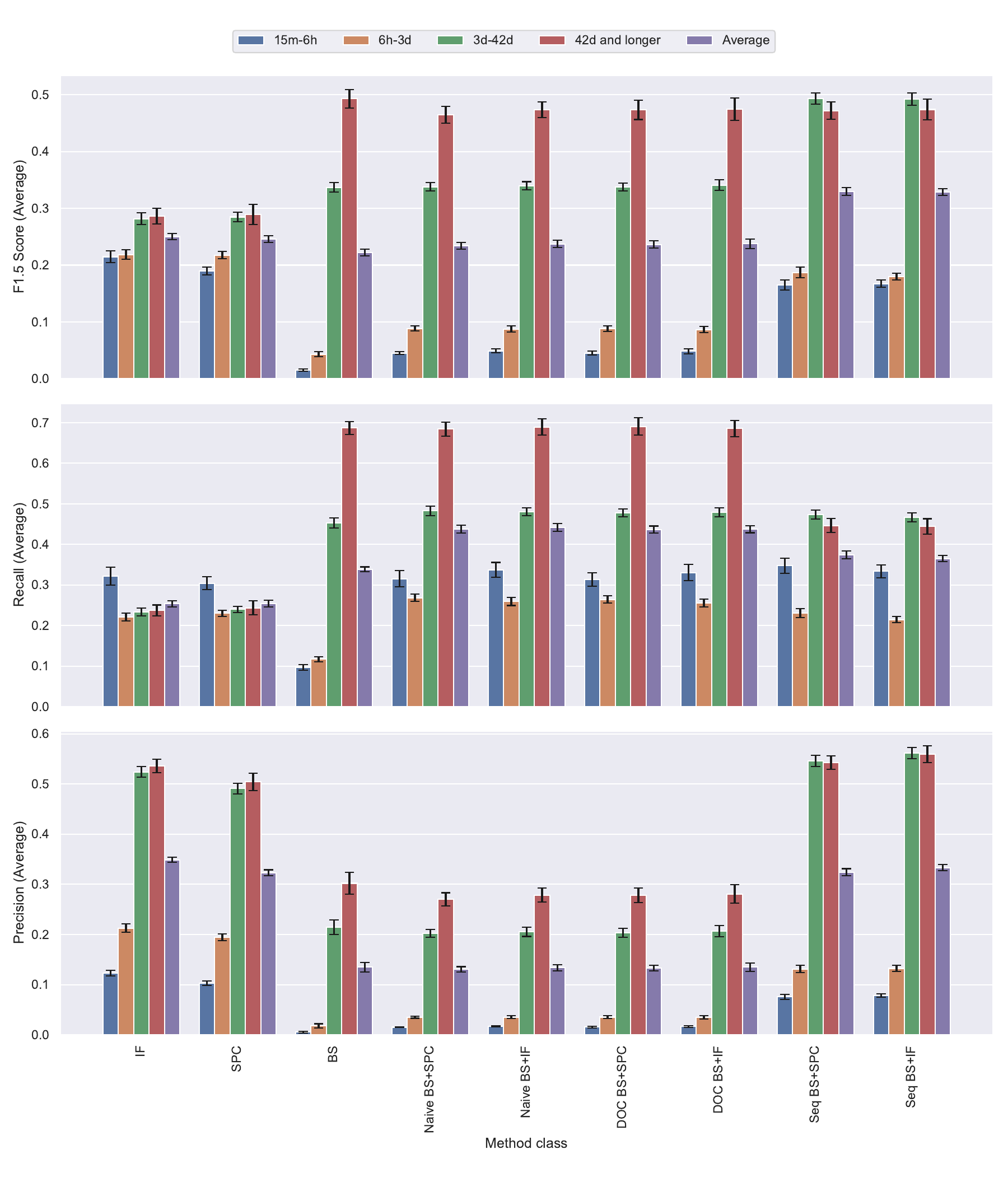}
    
\caption{From top to bottom: bar plots of the results of each method per length category for the F1.5, recall, and precision respectively. The height of each bar indicates the average score over the bootstrap iterations. The error bars indicate the standard deviation resulting from the bootstrap resampling.}
\label{fig:bootstrap_results_per_category}
\end{figure}

After evaluating each method on the validation set, we have selected for each method and ensembling combination the model with the best performance on the validation set. Each of these models was then evaluated on the test set to get an estimate of how well optimized models perform on unseen data. The resulting performance in terms of the F$1.5$ is visualized in Figure~\ref{fig:bootstrap_results_per_category}. From this figure we can observe the performance of the three base learners, as well as the effectiveness of the ensembling strategies when applied to different combinations. 

We find that IF and SPC perform similarly across the board. This is partially to be expected, as they are designed to find short anomalies/events that are rare compared to normal data. IF, however is able to more easily model multimodal distributions and detect anomalies in the presence of multimodality. Since IF does not noticably outperform SPC, it seems that it can not leverage this advantage. This is further corroborated by visualization of typical scaled data, which is mostly unimodal. both IF and SPC are able to detect short anomalies/events better than binary segmentation, achieving a F$1.5$ score of approximately $0.2$ for both the ``15 minutes to 6 hours", and ``6 hours to 3 days" event length categories. Perhaps surprisingly, these methods are able to detect some of the longer switch events as well, achieving just below $0.3$ F$1.5$ scores. The relatively good performance on these longer events can be easily explained. Many of the longer switch events have a larger distance to the $\vect{\delta}$ median than normal data. SPC is therefore able to detect these events based on this distance, while IF identifies them because these switch events occupy a lower density region where they are not masked by normal data. We find that even when ensembled in any of the three tested manners, IF offers no significant performance boost over SPC. Due to the ease of interpretability of SPC, it is preferable over IF.

Binary segmentation is found to be much better at detecting longer events than either SPC or IF. The longer the event, the better binary segmentation will perform, as can be seen from the performance on the longest category, which nears a F$1.5$ score of 0.5. It is also clear that binary segmentation fails to detect most short events/anomalies, achieving near random baseline F$1.5$ scores. 

The results from the naive and different optimization criterion (DOC) ensembles are less obvious. Both methods perform similarly, and the different optimization criterion addition does not significantly improve overall model performance. Perhaps surprisingly, the F$1.5$ score for the two shorter event length categories is only marginally better than that of binary segmentation. Indeed, ensembling binary segmentation with a detector for shorter events barely increases the F$1.5$ score. This can be explained when we consider the two terms making up the F$\beta$ score: precision and recall. When ensembling using an OR operation, we expect the recall to always increase, as we will only classify more samples in category ``1". The precision, however, will generally decrease, as more false positives will be found. Indeed, when looking at the performance in terms of recall and precision, which we have visualized in Figure~\ref{fig:bootstrap_results_per_category}, we can readily observe that the recall only increases or stays the same when ensembling. From this we conclude that the number of false positives introduced by OR ensembling causes simple ensemble methods to underperform.

Sequential ensembles seemingly do not suffer from this increase in false positives as much. We can see that sequential ensembles nearly match the performance of the individual SPC and IF detectors in the single categories where they perform best, while the performance in the ``3 to 42 days" category is significantly better than any base detector or other ensemble. The performance in the longest category is comparable to that of the other ensembles. Sequential ensembles indeed outperform any base detector or other ensembling method. While intuitively one might think the performance increase in the shorter categories should be similar to other ensembling strategies, it seems that due to the sequential nature, we can optimize more on precision than is the case in other ensembles. This can also be observed in Figure~\ref{fig:bootstrap_results_per_category}, where we see that the recall of the sequential models is lower in the longest category, while being similar in the other three. This means that all performance increases from naive or DOC ensembles to sequential can be attributed to a higher precision, which is confirmed by Figure~\ref{fig:bootstrap_results_per_category}.

\begin{figure}
\includegraphics[width=\textwidth]{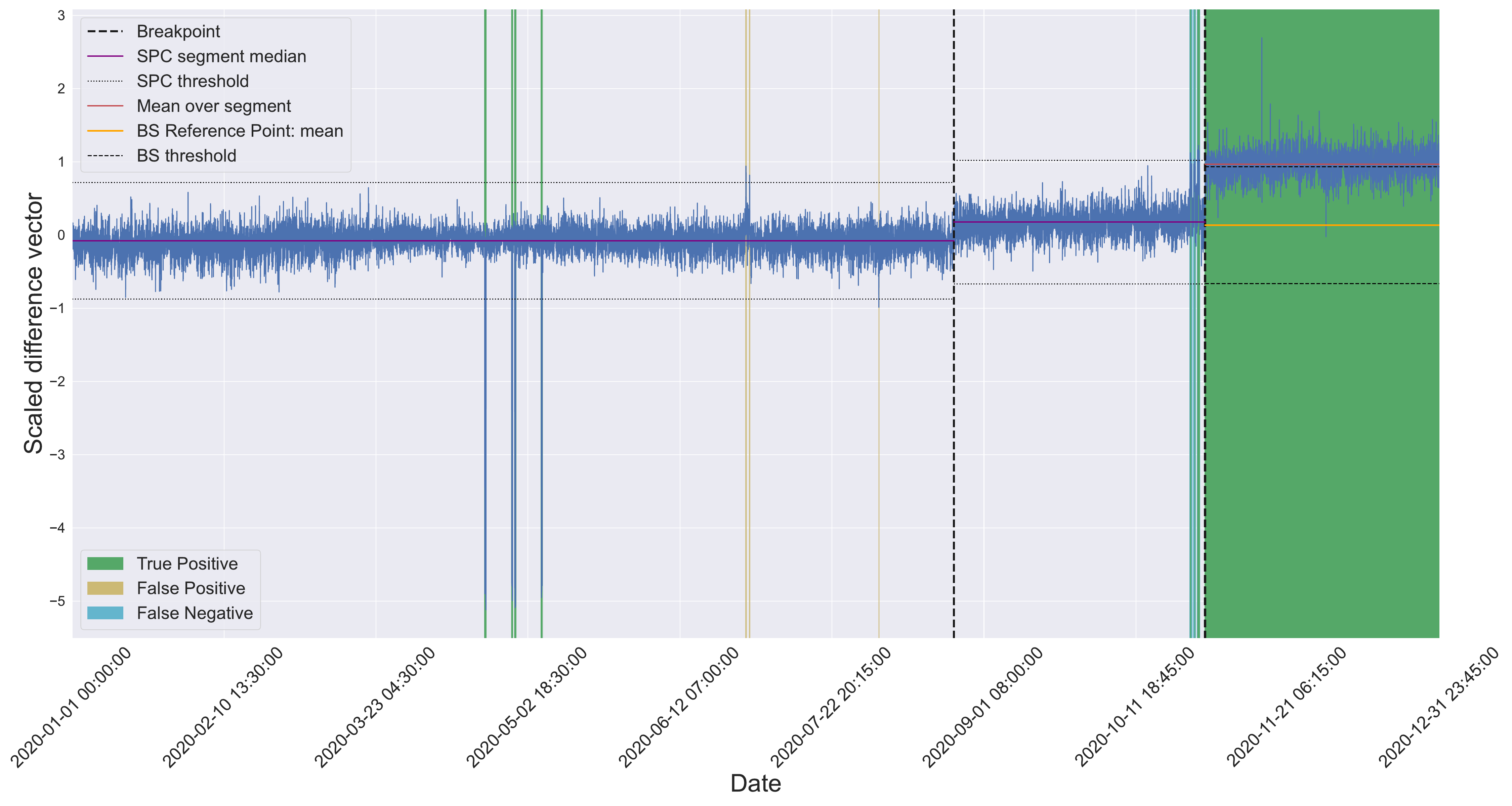}
\caption{Plot of the results of the best sequential BS+SPC model on station ``042" contained in the test set. The blue line indicates the inputted difference vector $\vect{\delta}$. The thresholds found by the initial BS pass are indicated by vertical dashed lines. The SPC segment medians and the BS overall mean used to calculate difference with the reference point are indicated by purple and orange lines. The boundaries for classification are indicated by either dotted or dashed lines for BS and SPC respectively. True positives, false positives and false negatives are visualized by changing the background color to green, yellow, or blue respectively.} 
\label{fig:sequential-model-example}
\end{figure}

To further study the behaviour of one of the sequential ensembles, specifically the combination of BS and SPC, we have visualized one of the station difference vectors together with the predictions, thresholds, and reference points from both components of the sequential ensemble in Figure~\ref{fig:sequential-model-example}. From this figure we can observe several qualities, as well as failings, of the sequential ensembling approach. Most prominent is the correct classification of the switch event on the far right of the figure. Binary segmentation was able to correctly classify this. Also correctly detected are the 3 shorter events/anomalies to the left of the figure, which SPC has detected due to their large negative contribution. However, some mistakes are made by the method. Specifically some false positives arise in the middle of the figure, where some points fall just outside the detection boundaries. Then, on the right in the second segment, there is a mix of true positives and false negatives. Due to the variability of the signal, only a few time points within this event are accurately detected by SPC. Lastly, one could argue that the second segment in its entirety should have been classified as a switch event. Yet, this was neither done by the domain expert, nor by the binary segmentation algorithm. Upon closer inspection, we found that the information in the load measurement and the bottom-up load is insufficient to fully determine whether this was a switch event. From this, and other observations on similar stations, we are led to conclude that further improvements can be made by including more metadata in future endeavours to improve our algorithms. This is further reinforced by the performance of all methods, which is relatively low across the board, even on the train data. This indicates that the problem is hard to learn, though it generalizes fairly well. Additionally, with these visualizations, we show that these models, even when ensembled, are exceedingly interpretable, solidifying their use in applications of societal importance.  

\begin{figure}
\includegraphics[width=\textwidth]{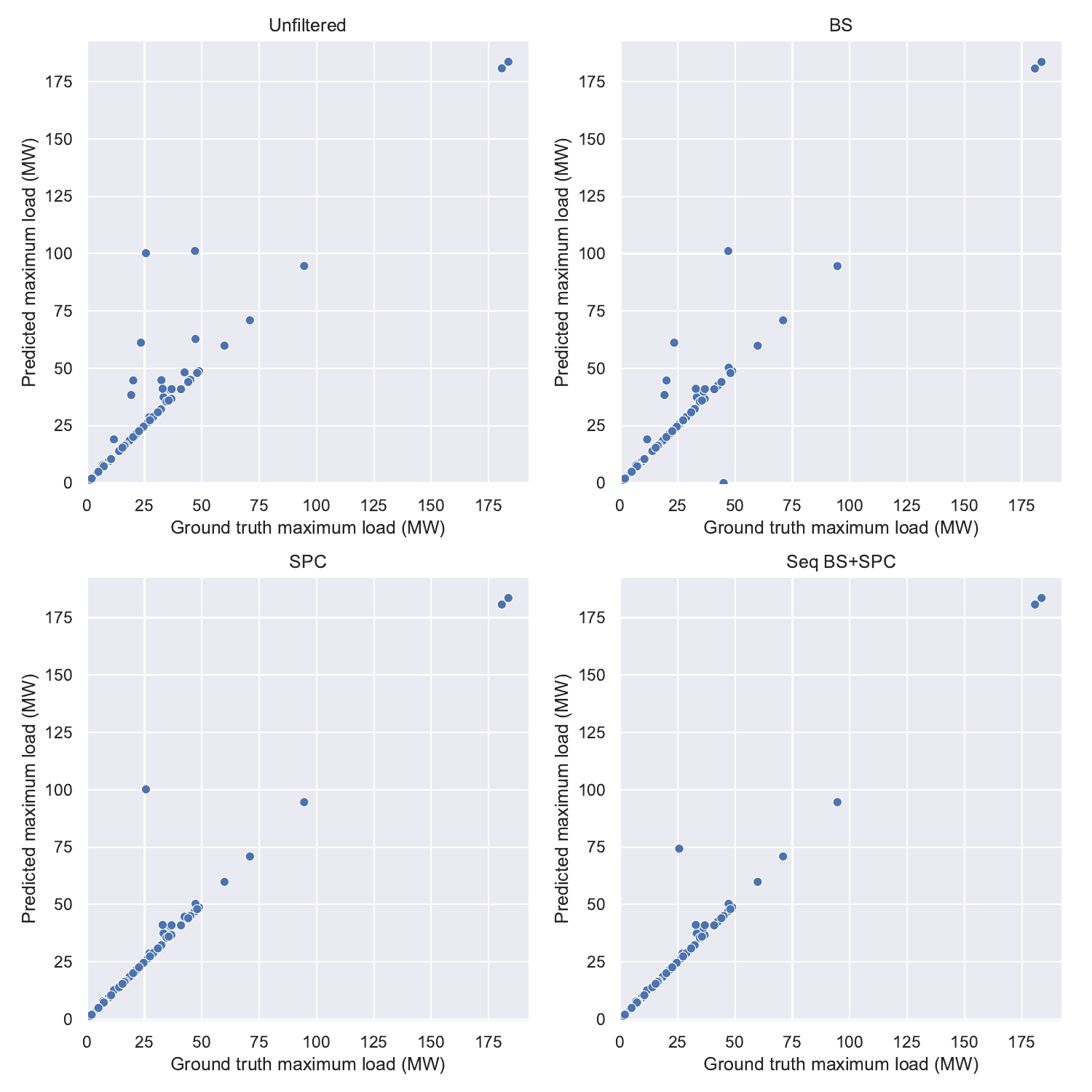}
\caption{Scatter plots where the ground truth maximum load estimate (in kW on x-axis) is plotted against the predicted maximum load estimate (in kW on y-axis). From top-left to bottom-right are shown the estimates resulting from: no filtering, the best Binary Segmentation (BS) model, the best Statistical Process Control (SPC) model, and the best Sequential Binary Segmentation + Statistical Process Control ensemble. When a point is above the $y=x$ line, too few points are filtered out, when the point is below, too many points are filtered out.} 
\label{fig:maximum_load_estimates}
\end{figure}

\begin{figure}
\includegraphics[width=\textwidth]{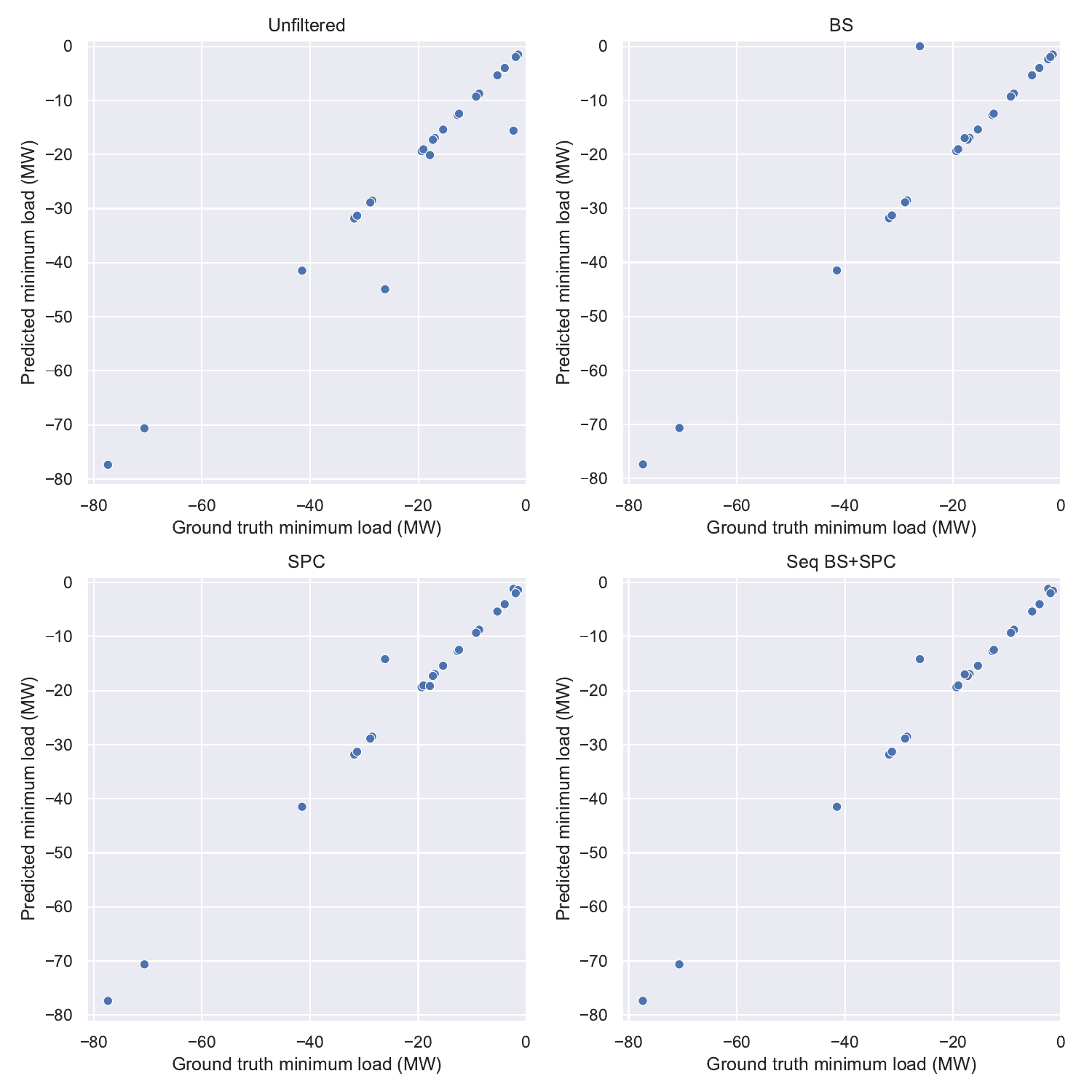}
\caption{Scatter plots where the ground truth minimum load estimate (in kW on x-axis) is plotted against the predicted minimum load estimate (in kW on y-axis). From top-left to bottom-right the figure shows the estimates resulting from: no filtering, the best Binary Segmentation (BS) model, the best Statistical Process Control (SPC) model, and the best Sequential Binary Segmentation + Statistical Process Control ensemble. Minimum load estimates are only shown for those stations that have measurements with a negative sign (23/60). When a point is above the $y=x$ line, too many points are filtered out, when the point is below, too few points are filtered out.} 
\label{fig:minimum_load_estimates}
\end{figure}

In order to gain more insight into how the proposed filtering approach works for automatically acquiring load estimates, we have plotted the ground truth load estimates against the predictions or unfiltered load estimates in Figure~\ref{fig:maximum_load_estimates} for the maximum load estimates, and in Figure~\ref{fig:minimum_load_estimates} for the minimum load estimates. In these figures, a perfect prediction would lead to all points lying on the diagonal. As can be seen, when we do not apply any filtering approach on the data, the minimum and maximum load values are highly inflated, leading to a 300\% increase at worst. A lot of potential grid capacity is unused when the measurements remain unfiltered and are being considered normal behavior. When we apply binary segmentation, we correctly predict some of the minimum and maximum loads, but still miss many while at the same time making the mistake of vastly underestimating the minimum and maximum load in a single case. With statistical process control the filtering procedure is much better, but we still have the worst case scenario of a 300\% increase in maximum load prediction. By combining both using sequential ensembling, we leverage the strengths of either method, reducing the worst case scenario to an approximately 200\% increase, and having only a single additional underestimation in the minimum load estimates. To further zoom in on the performance, we can see how well the method performs within certain error margins. The maximum load predictions are perfect in 75.00\%, and within a 10\% margin in 88.33\% of all cases. The minimum load predictions are perfect in 86.96\% and within a 10\% error margin in 91.30\% of all cases.

Due to the high performance of our ensemble method for load estimation it has been adopted for use within Alliander. Previously, all filtering and load estimates were done by hand. This was a time-consuming process, taking a month of time for several full-time employees. This was done once a year to acquire the load estimates for future planning and operations management. Earlier variations of the presented methodology have been in use since 2021, fully replacing the manual, time-consuming, process. In addition to replacing time-consuming manual labor of domain experts whose expertise is better utilized elsewhere, our methodology allows for easier updating of load estimates, which can now be generated on-the-fly. We hope that by open sourcing our methodology, code, and data other distribution system operators can benefit similarly.

\section{Discussion}

This paper showcases a novel combination of change point detection and anomaly detection algorithms for acquiring better load estimates. While the considered algorithms demonstrate good performance, we envision several possible improvements of the proposed method in the future

Firstly, in some cases, just looking at the difference between load measurement and bottom-up load is not enough to determine the ground truth. Additional information or metadata is needed for accurate segmentation, for example one could use a separate detector for seeing whether the load or bottom-up is incorrect, and incorporating this as a rule-based detector. We estimate that we could improve performance by using additional rule-based detectors on top of our current method.

Secondly, the datasets have a somewhat skewed distribution of events, meaning not all categories are equally represented in each dataset, as can be glanced from Table~\ref{table:event_length_stats}.  Due to this imbalance, as well as the general heterogeneity of events, we observe that the datasets are not as indistinct as they ideally would be. By labeling more stations we might be able to alleviate this problem. As labeling data is costly, the gains of this procedure might however not be worth the investment. Specifically over 500 primary substations are measured, which is unfeasible to manually label. Furthermore, events and anomalies are rare, and only a finite number of these occurs during any given year, which means that data acquisition and labeling for several more years might be needed to get fully balanced data. 

Our current method relies heavily on the bottom-up loads which are used to attain the difference vector $\vect{\delta}$. The accuracy of the bottom-up load varies between stations, leading to a high variability between stations in terms of variance. We correct for this using the robust scaling procedure, but it remains a critical step in the total analysis pipeline. When the bottom-up load is not available, for example due to anomalous situations in the grid, or when wanting to apply this method on old historical data, our methodology cannot be directly applied.

The current analysis considers one year of measurements for one station at a time. This is in line with the labeling procedure used by the grid operator prior to this research, where the data was labeled by hand during the first month of the year. As load estimates are used in long-term decision making, generally the speed of acquisition is not that important. However, should the availability of more recent load estimates be prioritized, the binary segmentation algorithm, which only works on static data, will need to be replaced by an online change point detection algorithm.

The methods employed for the change point and anomaly detection are all highly interpretable and fairly easily comprehensible. In recent years, machine learning has been widely applied in high-performance applications on time series. Examples are LSTM's for anomaly detection~\citep{lindemann2021survey}, and Meta's (formerly Facebook's) Prophet~\citep{taylor2018forecasting} for simultaneous anomaly and change point detection. While promising, these methods would not be readily usable on this application. These methods require far more data than the unsupervised methods used in this research, something which is unfeasible due to time of data acquisition. While these methods can be trained on the unlabeled time series, we still need labels for hyperparameter optimization. Using the model with the best predictions does not guarantee the best model for event detection, as an overparametrized model might learn to predict anomalies. Furthermore it has been shown that in many cases simple solutions work better than complex ones for time series anomaly detection~\citep{wu2021current}.

We show that our automatic filtering and load estimate procedure works in most cases, but there are still some failures cases. Future work could include studying these failure cases and devising detection methods for this purpose. Increasing the amount of data is expected to further increase performance. Furthermore, incorporating information of previous yearly measurements might lead to additional robustness of the filtering procedure. For example, a rule-based system could detect large differences in load estimates between years on top of the presented models, while still allowing for the full interpretability. 

It should be noted that while figures like Figure~\ref{fig:sequential-model-example} can be interpreted fairly easily, it is not immediately clear why the thresholds are chosen like they are. One could change the thresholds for this figure and achieve a better filtering. When interpreting these figures, one should always note that the thresholds are optimized to find the best F$1.5$ score over all stations, rather than the single figure shown.

\section{Conclusion}
Based on our research we present interpretable methodology for the automatic filtering of anomalies and switch events from load measurements in order to establish more accurate load estimates.

We posit that using unsupervised methods, with supervised optimization of hyperparameters and the threshold parameters, based on the F$1.5$ score, allows for robust, well-generalizing, performance on unseen data. 

We show that without filtering, a lot of grid capacity is left unused. In our experiments on unseen test data, comprised of 60 individual station measurements, we only observe a single severe failure case in both the automatic minimum and maximum load estimate predictions. Of all estimated predictions, approximately 90\% fall within a 10\% error margin. 

By having compared different methods and ensembling strategies we find that a combination of the well-known binary segmentation algorithm for change point detection and the tried statistical process control method for anomaly detection works best. The best ensembling strategy is a sequential ensemble, where the anomaly detector is applied after first segmenting the time series based on the obtained change points. The proposed methods are highly interpretable, a distinct advantage when this methodology is used in critical infrastructure planning. This high interpretability is a direct result of each underlying model being interpretable. SPC and BS are both simple, yet effective mathematical models. The strategies used for optimization are similarly straightforward, and can be visualized. Depending on underlying needs or when business priorities change the chosen threshold(s) for the algorithms can be adjusted based on the easily translatable precision and recall measures using figures such as Figure~\ref{fig:threshold_optimization}.

We finally identify possible steps for further improvement of the presented methodology. Incorporating additional data, either for further optimization, or as a historical reference, are potential avenues for improvement. Furthermore, the currently identified failure modes might be caught by using interpretable rule-based classification without losing the initial performance of the current algorithms.

\section*{Acknowledgements}
The research reported in this paper has been partly funded by the NWO grant NWA.1160.18.238 (\href{https://primavera-project.com/}{PrimaVera}); as well as BMK, BMDW, and the State of Upper
Austria in the frame of the SCCH competence center INTEGRATE [(FFG grant no. 892418)] part of the FFG COMET Competence Centers for Excellent Technologies Programme; the Alliander Research Centre for Digital Technologies; and finally by a Radboud Interdisciplinary Research Platform Green IT voucher.
We also want to thank Ayan Wasame for her efforts in studying label quality, and Evander van Wolfswinkel, Gijs van Paridon and Jari Immerzeel for their efforts in (re)labeling the data.


\bibliographystyle{elsarticle-num-names} 
\bibliography{refs.bib}





\clearpage
\appendix

\section{Evaluated and Best Hyperparameters}
\label{appendix:hyperparameters}
An overview of all evaluated hyperparameters can be found in Table~\ref{table:evaluated_hyperparameters}. The best parameters resulting from optimization on the validation set for each method-ensembling method combination can be found in Table~\ref{table:best_hyperparameters_first_half} and Table~\ref{table:best_hyperparameters_second_half}.

\begin{table*}
\caption{Evaluated hyperparameters for each method.}

\label{table:evaluated_hyperparameters}
\centering

\resizebox{\columnwidth}{!}{%
\begin{tabular}{lll}
\toprule
   &                    &                           Hyperparameter values \\
Method & Hyperparameter &                                                 \\
\midrule
\multirow{2}{*}{IF per station} & $n_{\textrm{estimators}}$ &                                          [1000] \\
   & Threshold strategy &                                   [Symmetrical] \\
\cline{1-3}
\multirow{3}{*}{IF over all stations} & $n_{\textrm{estimators}}$ &                                          [1000] \\
   & ($q_{\textrm{lower}}\%$, $q_{\textrm{upper}}\%$) &                  [(10, 90), (15, 85), (20, 80)] \\
   & Threshold strategy &                                   [Symmetrical] \\
\cline{1-3}
\multirow{2}{*}{SPC} & ($q_{\textrm{lower}}\%$, $q_{\textrm{upper}}\%$) &                  [(10, 90), (15, 85), (20, 80)] \\
   & Threshold strategy &                     [Symmetrical, Asymmetrical] \\
\cline{1-3}
\multirow{7}{*}{BS} & $\beta$ &         [0.005, 0.008, 0.015, 0.05, 0.08, 0.12] \\
   & $l$ &                                 [150, 200, 288] \\
   & $j$ &                                         [5, 10] \\
   & ($q_{\textrm{lower}}\%$, $q_{\textrm{upper}}\%$) &                  [(10, 90), (15, 85), (20, 80)] \\
   & $C$ &                                            [L1] \\
   & $\textit{reference\_point}$ &  [mean, median, longest\_median, longest\_mean] \\
   & Threshold strategy &                     [Symmetrical, Asymmetrical] \\
\bottomrule
\end{tabular}

}
\end{table*}

\begin{table*}
\caption{The best parameters resulting from optimization on the validation set for each single method and within the naive ensembles.}

\label{table:best_hyperparameters_first_half}
\centering

\resizebox{\columnwidth}{!}{%
\begin{tabular}{lllll}
\toprule
 &  &  &  & Hyperparameter values \\
Ensemble method & Combination & Method & Hyperparameter &  \\
\midrule
\multirow[t]{15}{*}{No ensemble} & \multirow[t]{15}{*}{-} & \multirow[t]{4}{*}{IF} & $n_{\textrm{estimators}}$ & 1000 \\
 &  &  & ($q_{\textrm{lower}}\%$, $q_{\textrm{upper}}\%$) & (15, 85) \\
 &  &  & Threshold strategy & Symmetrical \\
 &  &  & Optimal threshold(s) & 1.265484 \\
\cline{3-5}
 &  & \multirow[t]{3}{*}{SPC} & ($q_{\textrm{lower}}\%$, $q_{\textrm{upper}}\%$) & (15, 85) \\
 &  &  & Threshold strategy & Symmetrical \\
 &  &  & Optimal threshold(s) & 2.496898 \\
\cline{3-5}
 &  & \multirow[t]{8}{*}{BS} & $\beta$ & 0.008000 \\
 &  &  & $j$ & 10 \\
 &  &  & $l$ & 200 \\
 &  &  & $C$ & L1 \\
 &  &  & ($q_{\textrm{lower}}\%$, $q_{\textrm{upper}}\%$) & (10, 90) \\
 &  &  & $\textit{reference\_point}$ & mean \\
 &  &  & Threshold strategy & Asymmetrical \\
 &  &  & Optimal threshold(s) & (-0.4082615619841653, 0.6558452085588331) \\
\cline{1-5} \cline{2-5} \cline{3-5}
\multirow[t]{23}{*}{Naive} & \multirow[t]{11}{*}{BS+SPC} & \multirow[t]{8}{*}{BS} & $\beta$ & 0.008000 \\
 &  &  & $j$ & 10 \\
 &  &  & $l$ & 200 \\
 &  &  & $C$ & L1 \\
 &  &  & ($q_{\textrm{lower}}\%$, $q_{\textrm{upper}}\%$) & (10, 90) \\
 &  &  & $\textit{reference\_point}$ & mean \\
 &  &  & Threshold strategy & Asymmetrical \\
 &  &  & Optimal threshold(s) & (-0.4082615619841653, 0.6558452085588331) \\
\cline{3-5}
 &  & \multirow[t]{3}{*}{SPC} & ($q_{\textrm{lower}}\%$, $q_{\textrm{upper}}\%$) & (15, 85) \\
 &  &  & Threshold strategy & Symmetrical \\
 &  &  & Optimal threshold(s) & 2.496898 \\
\cline{2-5} \cline{3-5}
 & \multirow[t]{12}{*}{BS+IF} & \multirow[t]{8}{*}{BS} & $\beta$ & 0.008000 \\
 &  &  & $j$ & 10 \\
 &  &  & $l$ & 200 \\
 &  &  & $C$ & L1 \\
 &  &  & ($q_{\textrm{lower}}\%$, $q_{\textrm{upper}}\%$) & (10, 90) \\
 &  &  & $\textit{reference\_point}$ & mean \\
 &  &  & Threshold strategy & Asymmetrical \\
 &  &  & Optimal threshold(s) & (-0.4082615619841653, 0.6558452085588331) \\
\cline{3-5}
 &  & \multirow[t]{4}{*}{IF} & $n_{\textrm{estimators}}$ & 1000 \\
 &  &  & ($q_{\textrm{lower}}\%$, $q_{\textrm{upper}}\%$) & (15, 85) \\
 &  &  & Threshold strategy & Symmetrical \\
 &  &  & Optimal threshold(s) & 1.265484 \\
\cline{1-5} \cline{2-5} \cline{3-5}
\bottomrule
\end{tabular}

}
\end{table*}

\begin{table*}
\caption{The best parameters resulting from optimization on the validation set for each method within the DOC and sequential ensembles.}

\label{table:best_hyperparameters_second_half}
\centering

\resizebox{\columnwidth}{!}{%
\begin{tabular}{lllll}
\toprule
 &  &  &  & Hyperparameter values \\
Ensemble method & Combination & Method & Hyperparameter &  \\
\midrule
\multirow[t]{23}{*}{DOC} & \multirow[t]{11}{*}{BS+SPC} & \multirow[t]{8}{*}{BS} & $\beta$ & 0.008000 \\
 &  &  & $j$ & 10 \\
 &  &  & $l$ & 200 \\
 &  &  & $C$ & L1 \\
 &  &  & ($q_{\textrm{lower}}\%$, $q_{\textrm{upper}}\%$) & (10, 90) \\
 &  &  & $\textit{reference\_point}$ & mean \\
 &  &  & Threshold strategy & Asymmetrical \\
 &  &  & Optimal threshold(s) & (-0.4082615619841653, 0.6558452085588331) \\
\cline{3-5}
 &  & \multirow[t]{3}{*}{SPC} & ($q_{\textrm{lower}}\%$, $q_{\textrm{upper}}\%$) & (15, 85) \\
 &  &  & Threshold strategy & Symmetrical \\
 &  &  & Optimal threshold(s) & 2.496898 \\
\cline{2-5} \cline{3-5}
 & \multirow[t]{12}{*}{BS+IF} & \multirow[t]{8}{*}{BS} & $\beta$ & 0.008000 \\
 &  &  & $j$ & 10 \\
 &  &  & $l$ & 200 \\
 &  &  & $C$ & L1 \\
 &  &  & ($q_{\textrm{lower}}\%$, $q_{\textrm{upper}}\%$) & (10, 90) \\
 &  &  & $\textit{reference\_point}$ & mean \\
 &  &  & Threshold strategy & Asymmetrical \\
 &  &  & Optimal threshold(s) & (-0.4082615619841653, 0.6558452085588331) \\
\cline{3-5}
 &  & \multirow[t]{4}{*}{IF} & $n_{\textrm{estimators}}$ & 1000 \\
 &  &  & ($q_{\textrm{lower}}\%$, $q_{\textrm{upper}}\%$) & (15, 85) \\
 &  &  & Threshold strategy & Symmetrical \\
 &  &  & Optimal threshold(s) & 1.265484 \\
\cline{1-5} \cline{2-5} \cline{3-5}
\multirow[t]{23}{*}{Sequential} & \multirow[t]{11}{*}{BS+SPC} & \multirow[t]{8}{*}{BS} & $\beta$ & 0.008000 \\
 &  &  & $j$ & 10 \\
 &  &  & $l$ & 200 \\
 &  &  & $C$ & L1 \\
 &  &  & ($q_{\textrm{lower}}\%$, $q_{\textrm{upper}}\%$) & (15, 85) \\
 &  &  & $\textit{reference\_point}$ & mean \\
 &  &  & Threshold strategy & Asymmetrical \\
 &  &  & Optimal threshold(s) & (-0.4888460867656923, 0.8424118235083808) \\
\cline{3-5}
 &  & \multirow[t]{3}{*}{SPC} & ($q_{\textrm{lower}}\%$, $q_{\textrm{upper}}\%$) & (10, 90) \\
 &  &  & Threshold strategy & Symmetrical \\
 &  &  & Optimal threshold(s) & 2.237353 \\
\cline{2-5} \cline{3-5}
 & \multirow[t]{12}{*}{BS+IF} & \multirow[t]{8}{*}{BS} & $\beta$ & 0.008000 \\
 &  &  & $j$ & 5 \\
 &  &  & $l$ & 150 \\
 &  &  & $C$ & L1 \\
 &  &  & ($q_{\textrm{lower}}\%$, $q_{\textrm{upper}}\%$) & (15, 85) \\
 &  &  & $\textit{reference\_point}$ & mean \\
 &  &  & Threshold strategy & Asymmetrical \\
 &  &  & Optimal threshold(s) & (-0.4888460867656923, 0.8447648680436677) \\
\cline{3-5}
 &  & \multirow[t]{4}{*}{IF} & $n_{\textrm{estimators}}$ & 1000 \\
 &  &  & ($q_{\textrm{lower}}\%$, $q_{\textrm{upper}}\%$) & (10, 90) \\
 &  &  & Threshold strategy & Symmetrical \\
 &  &  & Optimal threshold(s) & 1.281501 \\
\cline{1-5} \cline{2-5} \cline{3-5}
\bottomrule
\end{tabular}

}
\end{table*}

\section{AUC-ROC performance of each method}
\label{appendix:AUROC}
\begin{figure}

\includegraphics[width=\linewidth]{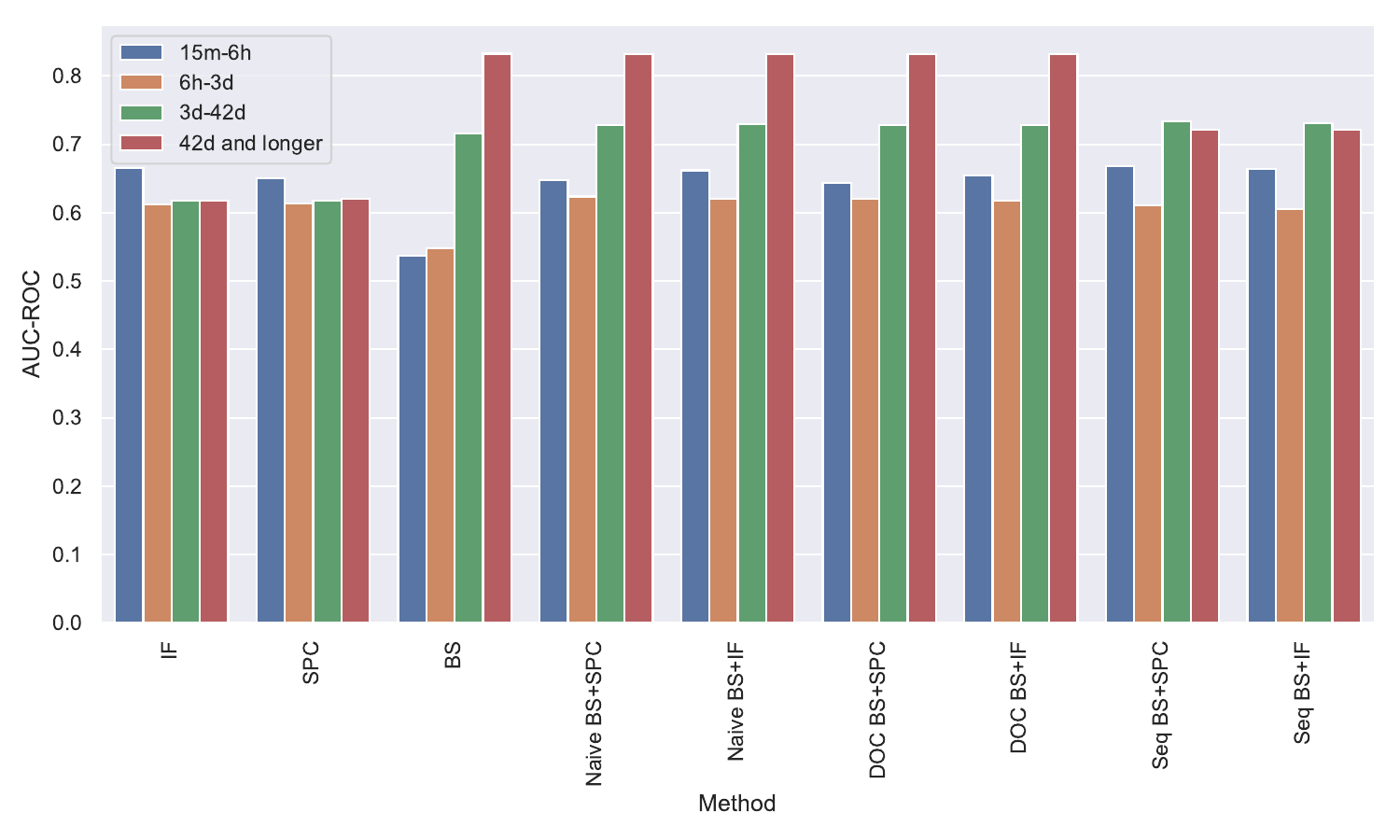}
    
\caption{A bar plot of the results of each method per length category in terms of the area under the curve of the receiver-operating characteristic  (AUC-ROC).}
\label{fig:auc_results_per_category}
\end{figure}

To provide insight into the ranking of the anomalies, so without explicit thresholding, we provide an additional plot of the area under the curve of the receiver-operating characteristic (AUC-ROC) in Figure~\ref{fig:auc_results_per_category}.

\end{document}